\documentclass[10pt]{article} %
\usepackage[preprint]{tmlr}

\usepackage{amsmath,amsfonts,bm}

\def\eqref#1{equation~\ref{#1}}

\def\1{\bm{1}}

\DeclareMathAlphabet{\mathsfit}{\encodingdefault}{\sfdefault}{m}{sl}
\SetMathAlphabet{\mathsfit}{bold}{\encodingdefault}{\sfdefault}{bx}{n}

\usepackage[dvipsnames,svgnames,x11names,table]{xcolor}
\usepackage[colorlinks=true, allcolors=RoyalBlue4]{hyperref}
\usepackage{url}
\usepackage{tabularx}
\usepackage{graphicx}
\usepackage{booktabs}
\usepackage{multirow}
\usepackage{makecell}
\usepackage[edges]{forest}
\usepackage{xcolor}
\usepackage{enumitem}
\usepackage{tcolorbox} %
\usepackage{hyperref}
\usepackage{booktabs}
\usepackage{caption}
\usepackage{subcaption}
\usepackage{wrapfig}
\providecolor{oc-gray-0}{HTML}{F8F9FA}
\providecolor{oc-gray-6}{HTML}{868E96}
\newtcolorbox[auto counter]{finding}{colback=oc-gray-0,colframe=oc-gray-6,
  boxrule=0.5pt,arc=2pt,left=5pt,right=5pt,top=3pt,bottom=3pt,
  before upper={\textbf{Finding~\thetcbcounter.}\ }}

\usepackage{mycommands}

\definecolor{forestgrhl}{RGB}{34, 139, 34}
\definecolor{figblue}{HTML}{2B7FB5}
\definecolor{figpink}{HTML}{D1597D}

\newcommand{\heart}{\ensuremath\heartsuit}

\title{Cognitive Foundations for Reasoning and\\Their Manifestation in LLMs}

\author{\name 
Shuyue Stella Li$^{1\heart}$,
Priyanka Kargupta$^{2\heart}$, \vspace{0.4mm}\\
Haocheng Wang$^{3\dagger}$, 
Jinu Lee$^{2\dagger}$, 
Dean Light$^{1\dagger}$, 
Shan Chen$^4$, 
Orevaoghene Ahia$^1$,
\vspace{0.4mm}\\
Thomas L. Griffiths$^3$,
Max Kleiman-Weiner$^1$,
Jiawei Han$^2$, 
Asli Celikyilmaz$^1$, 
Yulia Tsvetkov$^1$ \vspace{1mm}\\
\addr $^1$University of Washington, $^2$University of Illinois Urbana-Champaign, $^3$Princeton University, $^4$Harvard University \vspace{1mm}\\
{$^\heart$}Equal contribution, order determined with a coin toss.\hspace{2mm}{$^\dagger$}Equal second authors.\vspace{3mm}\\
}

\newcommand{\new}[1]{{#1}} 
\newcommand{\newtwo}[1]{{#1}}
\newcommand{\newthree}[1]{{#1}}
\newcommand{\newtom}[1]{{#1}}

\def\month{MM}  %
\def\year{YYYY} %

\begin{document}

\maketitle

\begin{abstract}
Large language models (LLMs) solve complex problems yet fail on simpler variants, suggesting they achieve correct outputs through mechanisms fundamentally different from human reasoning. 
To understand this gap, we synthesize cognitive science research into a taxonomy of 28 cognitive elements spanning reasoning invariants, meta-cognitive controls, representations for organizing reasoning and knowledge, and transformation operations. 
\new{We introduce a fine-grained evaluation framework and apply it to 192K traces from 18 models across text, vision, and audio, alongside 1,346 human think-aloud traces on a new, domain-agnostic dataset for human-model comparison.}
We find that models are able to deploy cognitive elements seen in human reasoning but differ in \emph{how} they use them. %
\new{Leveraging these patterns, we develop test-time reasoning guidance that automatically scaffolds successful reasoning topologies, improving the accuracy of the strongest models by \newtwo{9.7} percentage points on ill-structured problems.}
Meta-analysis of 1.6K LLM reasoning papers reveals the research community concentrates on easily quantifiable elements 
but \newthree{rarely targets self-awareness, whose use differs between human and model reasoning}. %
By establishing a shared vocabulary between cognitive science and LLM research, our framework enables diagnosis of reasoning failures and principled development of models that reason through robust cognitive mechanisms rather than spurious shortcuts, while providing tools to test theories of human cognition at scale.
\end{abstract}

\tableofcontents

\section{Introduction}

\new{Humans extrapolate from existing knowledge to unfamiliar scenarios and generate new knowledge, a process that constitutes reasoning \citep{lombrozo2024learning}.
Large language models (LLMs), by contrast, master complex skills \citep[such as proving Navier-Stokes blowup;][]{openai2026navierstokes,chervonyi2025gold,jiang2024survey} while failing at simple variants \citep[such as multiplication;][]{dziri2023faith,mccoy2024embers,mancoridis2025potemkin,berglund2023reversal,li2025personalized}.
This suggests that models reach correct outputs through mechanisms fundamentally different from the robust cognitive structures that let even young children flexibly deploy new concepts in unfamiliar domains \citep{bloom1998capacities}.
Understanding why is difficult, because training and evaluation reward outcomes without examining the processes that produce them \citep{lambert2024tulu3} and cannot distinguish genuine reasoning from memorization \citep{shao2025spurious,wu2025reasoning}. As a result, the field lacks both a conceptual framework for which cognitive elements should appear in model reasoning and empirical methods for measuring whether they do.}

\begin{figure}
    \centering
    \includegraphics[width=1.0\linewidth]{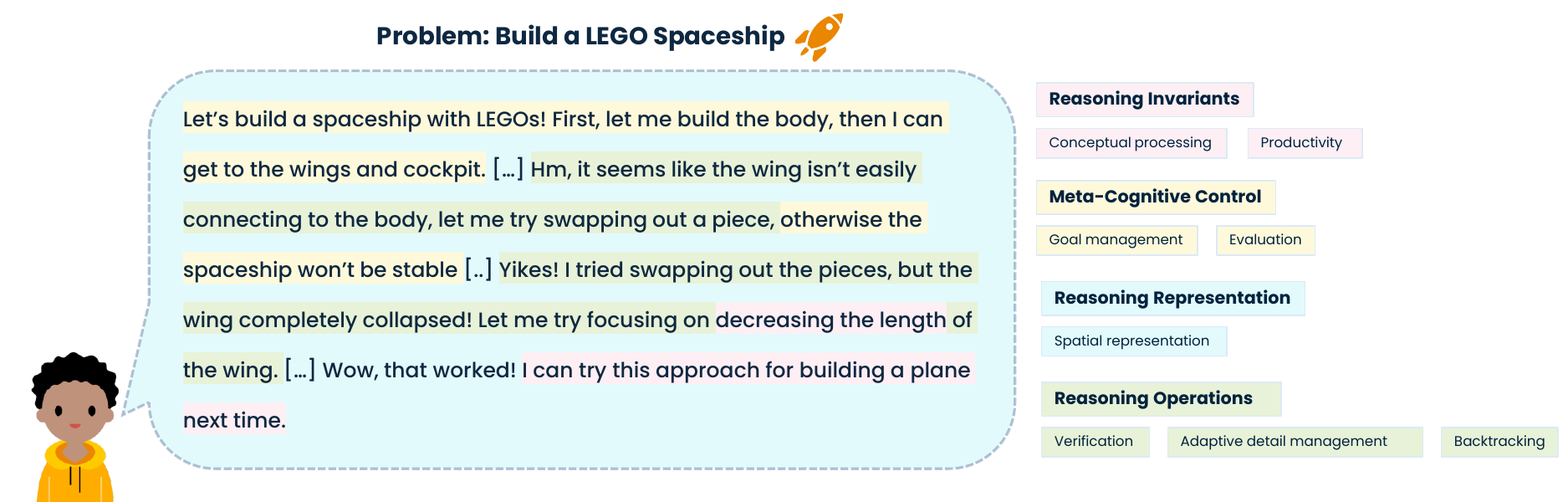}\vspace{-1mm}
    \caption{An example of cognitive elements present in a reasoning trace for building a LEGO spaceship. We characterize the different elements along the four dimensions of our taxonomy, as shown in Table \ref{tab:taxonomy}.}
    \label{fig:lego_example}\vspace{-6mm}
\end{figure}

Consider a child building a LEGO spaceship (Figure~\ref{fig:lego_example}). The child sets goals, decomposes the task, verifies connections, monitors progress, and reformulates failed approaches, orchestrating many cognitive processes into one coherent chain of thought. This orchestration is characteristic of human reasoning.
Research on LLM reasoning, by contrast, studies such behaviors one at a time, such as verification \citep{gandhi2025cognitive}, decomposition \citep{xu2025large}, and self-monitoring \citep{marjanovic2025deepseek}.
Our meta-analysis of 1,598 LLM reasoning papers (Section~\ref{sec:research_design}) shows this concentration on easily quantifiable behaviors: \newthree{46\%} examine sequential organization (e.g., step-by-step reasoning), while only \newthree{9\%} study backward chaining (e.g., working backwards from the answer).
Human reasoning research instead spans analogical transfer \citep{gentner1983structure,holyoak1989analogical}, causal inference \citep{sloman2009causal,gopnik2004theory}, representational flexibility \citep{ohlsson1992information,knoblich1999constraint}, and meta-cognitive control \citep{nelson1990metamemory,flavell1979metacognition}, which underlie navigating uncertainty, integrating knowledge, and adapting to novel situations. This provides a more integrated understanding of how cognitive elements interact to support flexible and generalizable reasoning. Thus, without a unifying framework between LLM and human reasoning, we risk optimizing for what we measure rather than what matters \citep{manheim2018categorizing}, mistaking narrow competence for broad reasoning ability.

To address this gap, we \textbf{introduce a unified taxonomy of cognitive foundations} that synthesizes theories of problem-solving, mental representation, and meta-cognition \citep[e.g.,][]{sweller1988cognitive,johnson1983mental,fodor1975language}.
We organize these into four dimensions: \textit{(1) \underline{Reasoning invariants}} are properties valid reasoning must maintain, such as logical coherence (not believing both ``the design is stable'' and ``the design will collapse'') and compositionality (understanding ``red LEGO cockpit'' by combining color, material, and function);
\textit{(2) \underline{Meta-cognitive controls}} select and monitor reasoning processes, such as context-awareness (recognizing missing LEGO pieces) and strategy selection (choosing between planning ahead and building through trial and error);
\textit{(3) \underline{Reasoning representations}} organize knowledge structurally and conceptually, such as hierarchical organization (decomposing the spaceship into its body, wings, and cockpit) and causal organization (understanding how weak support causes collapse);
\textit{(4) \underline{Reasoning operations}} construct, evaluate, and transform these representations, such as verification (checking whether LEGO pieces are securely connected), backtracking (revisiting earlier steps when a wing fails), and abstraction (recognizing that heavier components need more support).
Following Marr's \citeyearpar{marr1982vision} levels of analysis, invariants specify the properties valid reasoning must have, and representations and operations how it is carried out.
Meta-cognitive controls operate at an intermediate resource-rational level \citep{griffiths2015rational}, determining which strategies to deploy.
Across these levels, we identify 28 cognitive elements synthesized from existing cognitive science literature (Table~\ref{tab:taxonomy}), providing a common analytical vocabulary to bridge and jointly study human cognitive science and machine reasoning.

With this taxonomy, we conduct the \textbf{first large-scale empirical comparison of cognitive elements in human and LLM reasoning} across diverse problem types \citep{jonassen2000toward}: 192K traces from 18 models in text, vision, and audio, and 1,346 human think-aloud traces on 30 problems the models also solve (Table~\ref{tab:datasets}).
We find that \textbf{\textit{models possess nearly every cognitive element but \newthree{differ from humans in meta-reasoning}}}, \newthree{the monitoring and control of one's own reasoning, including when elements are deployed and how their outcomes are checked \citep{ackerman2017meta,russell1991principles}}.
Model traces reveal \newthree{failure} less clearly, where incorrect traces are still highly coherent, whereas incorrect human traces \newthree{become} incoherent and contain more self-awareness.
As problems become ill-defined and non-verifiable, %
humans and models both draw on fewer elements, although they exhibit opposite trends in self-monitoring and \newthree{meta-cognitive controls}. 
We also devise a novel \textbf{reasoning \newtwo{topology} representation} and find that models most often begin constructing a solution before representing the problem, whereas successful reasoning in both groups first identifies the problem's relevant features and constraints, as human reasoning most often does on \newtwo{diagnosis-solution and case analysis} problems.
These findings suggest that LLM reasoning could be brittle because models apply the procedures they have without adjusting when to use them to the problem at hand, which points to \textbf{\newthree{meta-reasoning} as a target for improving generalization}.

\new{Building on these patterns, we introduce \textbf{test-time reasoning guidance}, which scaffolds the \newtwo{reasoning topologies} of successful traces.
On dilemmas, for instance, successful traces first isolate the competing positions, then generate alternative resolutions, and then trace where each one leads (\textit{``which values are in conflict here?''} $\rightarrow$ \textit{``what are the possible ways to resolve the conflict?''} $\rightarrow$ \textit{``what follows from each option?''}), so we instruct the model to follow this \newtwo{topology}.
For the strongest models, guidance improves accuracy by \newtwo{9.7} percentage points on ill-structured problems and by 18.8 points on dilemmas, while weaker models lose accuracy. Capable models thus possess reasoning capabilities they do not deploy on their own, and understanding cognitive patterns can inform model capability improvements.}
Our contributions include:

\begin{enumerate}[itemsep=0pt,topsep=-3pt,leftmargin=16pt]
    \item A unified \textbf{taxonomy of cognitive foundations} synthesized from cognitive science research, containing 28 cognitive elements spanning reasoning invariants, meta-cognitive controls, reasoning representations, and reasoning operations, providing a unified vocabulary for characterizing reasoning beyond performance metrics.
    \item The first \textbf{large-scale empirical analysis} of 192K model traces and 1,346 human think-aloud traces, finding that models possess nearly every element but \newthree{differ from humans in meta-reasoning}, as their traces reveal failure less clearly, contain less self-monitoring and meta-cognitive controls on ill-structured problems, and most often begin constructing a solution before representing the problem.
    \item \textbf{Test-time reasoning guidance} that scaffolds empirically derived \newthree{topologies} improves the accuracy of the strongest models by \newtwo{9.7} percentage points on ill-structured problems, showing that capable models possess latent capabilities they fail to deploy without explicit structural guidance.
    \item \textbf{Meta-analysis of LLM research trends} over 1,598 LLM reasoning papers, showing that research rarely targets \newthree{self-awareness and goal management, two meta-cognitive controls whose use differs between human and model reasoning}.
\end{enumerate}

\section{Formalizing Cognitive Foundations for Reasoning}\label{sec:taxonomy}

\begin{table*}[h!]
\caption{Taxonomy of cognitive elements, organized along four main dimensions: Reasoning Invariants, Meta-cognitive Controls, Reasoning Representations, and Reasoning Operations.
}\label{tab:taxonomy}\vspace{-1.5mm}
\centering
\scriptsize
\renewcommand{\arraystretch}{1.4}
\setlength{\arrayrulewidth}{0.4pt}
\begin{tabularx}{\textwidth}{
|>{\raggedright\arraybackslash}p{0.17\textwidth}
|>{\raggedright\arraybackslash}p{0.245\textwidth}
|>{\raggedright\arraybackslash}X|}
\hline
\rowcolor{oc-pink-0}\multicolumn{3}{|l|}{\textbf{\textit{\hspace{-1mm}A. Reasoning Invariants:}} \textit{\new{Properties that valid reasoning maintains across steps, problems, and domains.}}} \\
\hline
\Invariant{Logical coherence} & \multicolumn{2}{l|}{Maintain consistency across reasoning steps and contexts \citep{fodor1988connectionism}.}  \\
\Invariant{Compositionality} & \multicolumn{2}{l|}{Build complex ideas from simpler components \citep{fodor1975language}.}  \\
\Invariant{Productivity} & \multicolumn{2}{l|}{Formulate an indefinite number of thoughts or solutions using a finite set of elements \citep{halford1989cognitive}.} \\
\Invariant{Conceptual processing} & \multicolumn{2}{l|}{\new{Operate over abstract concepts and relations, beyond surface and linguistic form} \citep{halford1989cognitive}.} \\[3pt]

\hline
\rowcolor{oc-yellow-0}\multicolumn{3}{|l|}{\textbf{\textit{\hspace{-1mm}B. Meta-Cognitive Controls:}} \textit{Higher-order abilities that select, monitor, and adapt processes.}} \\
\hline
\Metacognitive{Self-awareness} & \multicolumn{2}{l|}{Assess own knowledge state, capabilities, and task solvability \citep{wicklund1979influence}.} \\
\Metacognitive{Context awareness} & \multicolumn{2}{l|}{Perceive, understand, and navigate one's circumstances (including other agents) \citep{frith2007social}.} \\
\Metacognitive{Strategy selection} & \multicolumn{2}{l|}{Choose \& explore reasoning approaches suited to task and domain demands \citep{lieder2017strategy}.} \\
\Metacognitive{Goal management} & \multicolumn{2}{l|}{Establish, maintain, and adjust goals throughout the reasoning process \citep{griffiths2019doing}.} \\
\Metacognitive{Evaluation} & \multicolumn{2}{l|}{\new{Assess the quality, efficiency, and progress of the reasoning process and adapt it} \citep{fleming2017self}.} \\

\hline
\rowcolor{oc-cyan-0}\multicolumn{3}{|l|}{\textbf{\textit{\hspace{-1mm}C. Reasoning Representations:}} \textit{The formats and organizational patterns used to encode and relate knowledge and steps.}} \\
\hline
\multirow{2}{*}{\textbf{Structural}}
&\Representation{Sequential organization} & Order reasoning steps where their order matters \citep{skinner1953}. \\
\multirow{2}{*}{\textbf{Organization}}
&\Representation{Hierarchical organization} & Nest concepts in parent–child relationships \citep{galanter1960plans}. \\
&\Representation{Network organization} & Link concepts through multiple relationship types \citep{quillan1966semantic}. \\[3pt]
\hline

&\Representation{Ordinal organization} & Arrange elements by relative order or rank \citep{stevens1946theory}. \\
\textbf{Conceptual}&\Representation{Causal organization} & Connect elements through cause–effect relations \citep{heider1958psychology}. \\
\textbf{Organization}&\Representation{Temporal organization} & Relate events by before–after relations \citep{zakay1997temporal}. \\
&\Representation{Spatial organization} & Structure elements by spatial relationships \citep{tolman1948cognitive}. \\[4pt]

\hline
\rowcolor{oc-lime-0}\multicolumn{3}{|l|}{\textbf{\textit{\hspace{-1mm}D. Reasoning Operations:}} \textit{Goal-directed actions that construct, evaluate, modify, and navigate reasoning representations.}} \\
\hline
\textbf{Representation} & \Operation{Context alignment} & Align to task and situational demands \citep{gick1980analogical}. \\
\textbf{Selection}& \Operation{Knowledge alignment} & Align to domain-specific structures \& relations \citep{chi1981categorization}. \\[3pt]
\hline
\textbf{Representation} & \multirow{2}{*}{\Operation{Verification}} & \multirow{2}{=}{Check whether individual steps and conclusions meet pre-determined criteria \citep{flavell1979metacognition}.} \\
\textbf{Evaluation}&  &  \\
\hline
\multirow{5}{*}{\textbf{Representation}}
& \Operation{Selective attention} & Focus on relevant details and filter noise \citep{broadbent1958perception}. \\
\multirow{5}{*}{\textbf{Modification}}
& \Operation{Adaptive detail management} & Adjust granularity based on task demands \citep{Rosch1978-ROSPOC-8}. \\
& \Operation{Decomposition and integration} & Divide problems and synthesize subsolutions \citep{newell1959report}. \\
& \Operation{Representational restructuring} & Reformulate problems for new insights \citep{Wertheimer1945}. \\
& \Operation{Pattern recognition} & Detect recurring structures across contexts \citep{Selfridge1959}. \\
& \Operation{Abstraction} & Generalize from specific cases \citep{hull1920quantitative}. \\[3pt]
\hline
\multirow{2}{*}{\textbf{Representation}}
& \Operation{Forward chaining} & Reason from known facts toward goals \citep{huys2012bonsai}. \\
\multirow{2}{*}{\textbf{Navigation}}
& \Operation{Backward chaining} & Work backward from goals to prerequisites \citep{park2017relative}. \\
& \Operation{Backtracking} & Revisit and correct prior reasoning paths \citep{Nilsson1971ProblemsolvingMI}. \\[4pt]

\hline

\end{tabularx}\vspace{-4mm}
\end{table*}

Understanding the extent to which LLMs reason requires a theory of what reasoning is. 
Decades of cognitive science research on problem solving, mental representation, and meta-cognition provide the foundation for such a theory. We trace this research chronologically and then organize it into a single framework.

\new{The cognitive revolution of the 1950s introduced a framework \newtom{based on viewing} the mind as an information-processing system \citep{gardner1985mind,miller2003cognitive}. Chomsky's \citeyearpar{chomsky1966cartesian} account of the creative use of language showed how a finite system can generate an unbounded number of novel expressions, and Newell and Simon's work on symbolic problem solving, later stated as the physical symbol system hypothesis, held that the manipulation of symbols provides the necessary and sufficient means for general intelligent action \citep{newell1976computer}. Fodor's \citeyearpar{fodor1975language} Language of Thought Hypothesis brought these ideas together, proposing that thinking is a computational process operating over compositional representations \citep{quilty2023best}. Complex thoughts are composed of simpler components combined by rules, so a finite set of conceptual primitives can yield infinitely many novel thoughts. We are not ``hapless tourists bound by a limited phrasebook of ideas'' \citep{frankland2020concepts}, but fluent speakers of an internal, compositional language. However, this view of the mind as a symbolic logic engine could not explain findings that emerged from psychology.}

\new{Research in cognitive psychology showed that human reasoning often departs from the norms of symbolic logic. In the selection task, people often fail to apply a conditional rule stated in abstract terms yet succeed when a logically equivalent rule is framed as a familiar social rule, a pattern known as the content effect \citet{wason1968reasoning}. \citet{tversky1974judgment} showed that people rely on heuristics that violate principles of logic and probability theory in predictable ways. These results motivated theories that sought to explain both the successes and the failures of human reasoning. Dual-process theory distinguished fast, intuitive, and error-prone processing from slow and deliberative reasoning \citep{evans2003two,evans2013dual,oktar2022deciding}. Mental models theory proposed that reasoning constructs and manipulates semantic simulations of the world \citep{johnson1983mental,johnson-laird2010}, which accounts for content effects that purely syntactic inference rules cannot. Connectionist accounts explained reasoning as an emergent capability of networks of simple, neuron-like units \citep{sep-connectionism,rumelhart1986parallel}. Bayesian approaches characterized what reasoning computes, recasting rationality as probabilistic inference under uncertainty, in which reasoners update their beliefs from new, ambiguous, and noisy data in ways that approximate Bayesian inference \citep{griffiths2024bayesian,jacobs2011bayesian,oaksford2009precis}.}

This theoretical diversity, while valuable for cognitive science, has left research on LLM reasoning without a common framework. \citet{gandhi2025cognitive} characterize reasoning through verification, subgoal setting, backtracking, and backward chaining, while \citet{lee2024reasoning} focus on logical coherence, compositionality, and productivity. Because each line of work examines a different subset of reasoning, models can appear capable under one framework yet fail under another \citep{berglund2023reversal, mccoy2024embers, dziri2023faith}, and researchers lack a shared vocabulary for diagnosing where and why reasoning breaks down.
The theories reviewed above conflict less than they appear to, because they largely address different questions. Fodor's compositionality and Bayesian rationality characterize properties that any adequate reasoner must have, mental models and connectionist networks characterize the representations and processes that carry reasoning out, and dual-process theory concerns when effortful deliberation is engaged. 
Marr's framework describes an information-processing system at three levels of analysis \citep{marr1982vision}, providing a principled way to relate such accounts.

We use Marr's levels to organize cognitive theories of reasoning into a single framework, grouping the \textit{elements of reasoning} that each theory contributes by the level of analysis they address.
The \textit{computational level} specifies what the system computes and why.
We call the elements at this level \textbf{reasoning invariants}, the properties that any valid reasoning must have, independent of how it is carried out \citep{fodor1975language, fodor1988connectionism, thagard2002coherence}.
The \textit{algorithmic level} specifies the representations a system uses and the procedures that operate over them to achieve the computational goal.
Its elements divide into \textbf{reasoning representations}, which encode knowledge in structured form, and \textbf{reasoning operations}, which transform it.
The \textit{implementation level} concerns physical instantiation in neural hardware or silicon, which we set aside. 
\citet{griffiths2015rational} identify an intermediate level between the computational and algorithmic levels, which specifies which algorithms a reasoner should deploy given task demands and limited cognitive resources.
We call the elements at this level \textbf{meta-cognitive controls}, which select, monitor, and adapt reasoning strategies as a problem is solved.
Table~\ref{tab:taxonomy} lists the 28 elements of the resulting framework in these four dimensions, together with the foundational work behind each element. 

Our starting example in Figure~\ref{fig:lego_example} illustrates how these dimensions interact: consider again a child playing with a pile of LEGO bricks and building something novel, like a ``spaceship.''
At the computational level, the invariants of compositionality and productivity constrain the solution, which must be a complex new structure built from a finite set of simple components \citep{newell1972human}. 
At the resource-rational intermediate level, meta-cognitive controls choose and oversee the approach \citep{lieder2017strategy,griffiths2015rational}, as the child decides whether to plan upfront or build exploratively, tracks progress, and judges whether the emerging structure matches their intent.
At the algorithmic level, representations and operations carry out the build, as the child forms a three-dimensional mental model of the spaceship, organizes it hierarchically into body, wings, and cockpit, and refines it through operations such as decomposition, verification, and backtracking.

The example shows that reasoning arises from the interplay of all four dimensions. The constraints at the computational level call for strategic choices about how to build, these choices control which representations the child constructs, and operations then transform those representations \citep{griffiths2015rational}.
In particular, each element contributes to success only when it is deployed at the right time and together with the right elements.
Assessing how well a reasoner uses an element therefore requires studying both its presence and its relation to other elements.

\subsection{Reasoning Invariants: \new{Properties of Valid Reasoning}}\vspace{-1mm}

\new{Reasoning invariants are properties that valid reasoning maintains across steps, problems, and domains.}
\new{The language of thought hypothesis \citep{fodor1975language,fodor1988connectionism} identifies three of them. Because valid reasoning manipulates structured representations by compositional rules, the thoughts it produces are logically coherent (free from contradiction), compositional (with complex meanings built from simpler parts), and productive (able to yield an unbounded number of novel inferences from finite primitives).}
\new{We add conceptual processing from research on cognitive capacity and conceptual structure \citep{halford1989cognitive,kholodnaya2016conceptual}, which requires reasoning to operate over abstract concepts and their relations beyond the surface and linguistic form of a problem. These four invariants constrain any valid solution.}

\Invariant{Logical coherence} maintains consistency across reasoning steps and contexts \citep{rips1983cognitive, thagard2002coherence}. In the LEGO example, if the child simultaneously believes ``this wing design is stable'' and ``this wing design will collapse,'' reasoning breaks down. The contradiction creates cognitive dissonance that must be resolved by revising the stability judgment, reinforcing the structure, or reconsidering the design \citep{festinger1962cognitive, harmon2019introduction}. This pressure to restore consistency---both individually and socially---shapes whether and how reasoners revise mental models and update inferences amid inconsistency or disagreement \citep{oktar2025beliefs,mercier2017enigma}. 

\new{\Invariant{Compositionality} is the property that the meaning of a complex representation is determined by the meanings of its parts and the way they are combined \citep{fodor1975language,fodor2001language,fodor1988connectionism,russin2024frege}.}
Understanding ``red cockpit with transparent dome'' requires combining the concepts of color, component type, and material while preserving the meaning of each, and its meaning follows predictably from the properties of its parts and how they connect.
This property increases the expressive power of the cognitive system, enabling complex meanings to be constructed from a finite vocabulary of concepts \citep{frankland2020concepts, piantadosi2011learning}.

\Invariant{Productivity} is the behavioral capacity to generate and understand an unbounded number of novel thoughts never previously encountered \citep{fodor1975language, fodor1988connectionism}. While compositionality is a structural property of representations, productivity is a property of the reasoning system: it must actively deploy compositional structure to produce and interpret expressions beyond its prior experience. This distinction is critical for evaluating LLMs---a model may learn compositional representations yet fail to deploy them productively when faced with novel combinations \citep{dziri2023faith, hauser2002faculty}. Having understood how to build a spaceship, a productive reasoner spontaneously applies the same structural principles to generate a plane or a castle without requiring explicit instruction for each new case.

\new{\Invariant{Conceptual processing} is the property that reasoning operates over abstract concepts and their relations \citep{halford1989cognitive,kholodnaya2016conceptual,baron2008}.}
\new{This property allows reasoning to generalize across problems that share deep structure but differ on the surface.}
When building the LEGO spaceship, the child reasons about stability, symmetry, and aerodynamic form---abstract properties that transcend specific brick configurations. They access deeper knowledge: structural relationships (how weight distribution affects balance), functional constraints (wings must attach securely to support the body), and design principles (symmetry creates visual appeal). This conceptual depth distinguishes genuine reasoning from shallow pattern matching. \new{The child understands why bilateral symmetry serves functional and aesthetic goals, which allows them to apply it to designs they have never seen.}

\new{The four invariants constrain one another, since logical coherence determines which compositional structures are valid, productivity depends on compositional structure to generate novel thoughts, and conceptual processing ensures that the structures being composed and generated are relations among concepts.}
\new{Invariants specify the properties that valid reasoning must have. The mechanisms that detect when these properties are violated, select strategies, and monitor progress toward a solution are meta-cognitive controls.}

\subsection{Meta-Cognitive Controls: Executive Control}\label{sec:taxonomy:controls}\vspace{-1mm}

\new{Meta-cognitive controls are the executive functions that select, monitor, and adapt reasoning processes \citep{fleming2024metacognition}. They operate at the resource-rational level, deciding which representations and operations to deploy given the task and the reasoner's limited time and effort.}

\new{\Metacognitive{Self-awareness} is the capacity to assess one's own knowledge state, capabilities, and the solvability of a task, and the other controls build on it \citep{neisser1988five, wicklund1979influence,leary2003evolution}. Rochat \citeyearpar{rochat2003five} describes self-awareness as ``arguably the most fundamental issue in psychology, from both a developmental and evolutionary perspective.'' In the LEGO example, the child asks ``Am I good at building spaceships?'' or ``Do I know what a spaceship looks like?'' The answers determine how much effort to invest and which strategies are feasible.}

\Metacognitive{Context awareness} perceives and responds to situational demands, environmental constraints, and the presence of other agents \citep{boyd2011cultural,
frith2007social, lei2023sociality}. In the LEGO example, context shapes the task. Building alone for fun permits flexible exploration. Building with a friend introduces social dynamics requiring cooperation and negotiation. Building in a timed contest with limited bricks demands prioritizing speed and efficiency over creative exploration. Context awareness determines which strategies are appropriate and which goals are worth pursuing \citep{milli2021rational}.

Given this awareness of self and context, the child engages in \Metacognitive{strategy selection}, choosing an approach suited to task demands \citep{lieder2017strategy,lieder2020resource, mata2011easy}. They might adopt an exploratory, bottom-up strategy, combining bricks and allowing the design to emerge organically. Alternatively, they could choose a planned, top-down strategy, first visualizing the spaceship mentally and then finding bricks to realize each component. With limited bricks, they might adopt a resource-first strategy, sorting pieces to assess availability before committing to a design. With time constraints, they might use a schema-based strategy, quickly replicating a remembered design.

\new{These strategic choices are accompanied by \Metacognitive{goal management}, the process of setting, sequencing, and adjusting goals throughout reasoning \citep{griffiths2019doing, dolan2013goals, cushman2015habitual}. The abstract goal ``build a spaceship'' decomposes into sub-goals such as constructing the main body, adding wings, building the cockpit, and seating the pilots. These goals change as the task unfolds and as the reasoner's assessment of self and context changes. While working on the wings, the child might spot a clear brick that would make a good windshield and pause the current task to insert a new sub-goal. Conversely, if attaching wheels proves too difficult, the child abandons this sub-goal to finish the rest of the ship.}

\new{Progress toward these goals is monitored through \Metacognitive{evaluation}, which assesses the quality, efficiency, and progress of the reasoning process and adjusts it as needed \citep{Yeung2012-pv,fleming2017self, stipek1992self}. The child continually asks whether the spaceship is coming together as intended, whether the current approach is too slow, and whether a different configuration would work better. Evaluation differs from the verification operation (Section~\ref{sec:taxonomy:operations}). Verification checks whether a particular step or conclusion is correct, such as whether two pieces actually connect, while evaluation judges whether the reasoning as a whole is on track and decides whether to persist with the current approach or change strategies and goals.}

\new{The five controls operate as a single executive system. Self-awareness and context awareness establish what the reasoner can do and what the situation demands, strategy selection and goal management turn this assessment into a plan, and evaluation monitors the plan and triggers adaptation when it falters. In the LEGO example, the child's judgment of their own skill and of the time available determines whether they plan upfront or build exploratively, the chosen strategy determines the sub-goals, and evaluation of the emerging structure determines when to revise.}
Controls govern which processes are deployed but do not specify the representational structures they operate over, which are defined at the algorithmic level.

\subsection{Reasoning Representations: Organizational Structures}\label{sec:taxonomy:representations}
\vspace{-1mm}

\new{At the algorithmic level, the effectiveness of reasoning depends on how knowledge is structured \citep{sweller1988cognitive, britton1982effects}. This focus on structure has roots in associationism and the early study of how ideas become linked in the mind \citep{james1890principles}. Cognitive load theory explains why structure matters, since working memory is limited and poorly structured information overwhelms it while well-structured information is processed efficiently \citep{sweller2011cognitive}. The representational formats proposed in cognitive science differ in how they structure knowledge. Production systems encode procedural knowledge as if-then rules and underlie early models of human problem solving \citep{newell1972human}. Prototype theory represents a category by its most typical member \citep{rosch1975cognitive, rosch1975family}, and exemplar theories represent it by stored instances of its members \citep{medin1978context}. Semantic networks model knowledge as concepts connected by typed relations, capturing both hierarchical and associative organization \citep{quillan1966semantic, collins1975spreading, steyvers2005large}. Frames and schemas encode stereotyped knowledge about common situations and supply default expectations that guide prediction and inference \citep{minsky1974framework, Bartlett1932}. Mental models represent the structural and causal relationships of the domain being reasoned about \citep{johnson1983mental}.}
\new{Two kinds of organization recur across these formats, and we use them to group the representations in our taxonomy. Structural organization concerns how elements are connected, and conceptual organization concerns the semantic relations that connect them. A semantic network, for example, combines hierarchical and network structure, while a mental model combines causal, spatial, and temporal relations.}

\textbf{Structural organization} specifies the architecture through which elements relate \citep{kemp2008discovery}. 
\new{The simplest form is \Representation{sequential organization}, which orders reasoning steps where their order matters \citep{skinner1953}. Following a recipe, starting a car, and deriving a proof all require carrying out steps in the correct order. Sequential organization concerns the order of the reasoning itself, whereas temporal organization, described below, concerns before--after relations among events in the problem. Many reasoning tasks demand richer connectivity than a single sequence provides \citep{lashley1951problem, rosenbaum2007problem}.}
\Representation{Hierarchical organization} nests concepts in parent-child relationships, enabling decomposition of complex wholes into manageable parts \citep{galanter1960plans,botvinick2009hierarchically,haupt2018hierarchical}. Biological taxonomies classify organisms into kingdom, phylum, class, and order; problem-solving decomposes complex tasks into manageable subtasks; and administrative structures organize authority and responsibility through nested levels. This parent--child structure appears across many domains of human knowledge.
Yet even hierarchies have limitations. Many domains resist strict tree structures because elements relate through multiple relationship types simultaneously. 
\Representation{Network organization} captures this richer connectivity \citep{quillan1966semantic, shafto2008inductive}. \new{In an ecosystem, organisms are linked by predation, competition, symbiosis, and energy flow, and a network represents all of these relations at once.} 
In the LEGO example, the child's mental model exhibits hierarchical decomposition of the spaceship into body, wings, and cockpit, while simultaneously maintaining network relationships that specify how components physically connect and structurally support each other.

\textbf{Conceptual organization} \new{specifies the semantic relationships that give representations their inferential power.} 
When the child reasons about alternative wing designs for their LEGO spaceship, \Representation{ordinal organization} allows ranking them from best to worst, most stable to least stable \citep{stevens1946theory}. 
But ranking alone provides no explanatory depth. \Representation{Causal organization} connects elements through cause-effect relations, revealing why one design outperforms another \citep{heider1958psychology, khemlani2014causal}. Inadequate wing reinforcement causes structural instability, which prevents stable flight. 
This causal chain guides both diagnosis (why did it fail?) and intervention (how to fix it?). 
\new{Causal reasoning interacts closely with \Representation{temporal organization}, which relates events and states in the problem by before--after relations \citep{zakay1997temporal, Bartlett1932,hoerl2019thinking}. The child anticipates that the ship will be lifted, tilted, and landed during play, and that a loosely attached wing will fall off once the ship is tilted.}
Temporal constraints shape planning, while causal understanding determines which temporal orderings make sense. 
Finally, \Representation{spatial organization} structures elements geometrically, capturing relationships like adjacency, containment, and orientation \citep{tolman1948cognitive, shepard1971mental, landau1993whence}. The child reasons about which pieces fit where, how rotating a component changes available attachment points, and how weight distribution affects balance \citep{shelton2022characterizing,cortesa2017characterizing}. 
\new{The child moves among these organizations as the task demands, using spatial organization to decide which pieces fit, causal organization to predict whether the structure will hold, and temporal organization to anticipate what will happen to it during play.}

\subsection{Reasoning Operations: Transformation Procedures}\label{sec:taxonomy:operations}\vspace{-1mm}

At the algorithmic level, reasoning unfolds through goal-directed procedures that construct new representations, evaluate existing ones, modify them adaptively, and navigate through them strategically \citep{mcclelland2010letting,johnson1983mental}. 
These operations are combined flexibly under meta-cognitive control and constrained by reasoning invariants.
While humans possess a large repertoire of reasoning operations, we focus on the four most fundamental clusters that characterize reasoning across domains: selection, evaluation, modification, and navigation.

\textbf{Representation selection} operations align structures to the demands of the task and domain, a process rarely conscious but profoundly shaping subsequent reasoning \citep{newell1972human, tversky1981framing}. 
\Operation{Context alignment} chooses organizational schemas fitting the problem \citep{gick1980analogical, gick1983schema, gentner1983structure}: temporal for historical explanation, causal for scientific reasoning, spatial for navigation. 
\Operation{Knowledge alignment} maps problems onto domain-specific schemas \citep{chase1973perception, chi1981categorization, schoenfeld2014mathematical}. 
A physician diagnosing symptoms activates medical taxonomies; an auto mechanic activates mechanical systems and failure modes. 
In the LEGO example, the child chooses to organize the parts of the spaceship spatially and draws on their intuitive knowledge of gravity and physics.

\textbf{Representation evaluation} verifies reasoning steps against predetermined criteria. 
\Operation{Verification} checks intermediate inferences for consistency, plausibility, and coherence with known facts \citep{flavell1979metacognition, goldstein2010general}.
The child might verify that two pieces actually connect as expected by testing the attachment, or recount the number of attachment points to confirm stability. 
Humans continuously verify their reasoning, though the sophistication and systematicity \citep{Simon1978IndividualDI} and how consciously verification is carried out \citep{schneider1977controlled,anderson1982acquisition} all vary with expertise. 

\textbf{Representation modification} adaptively changes representational form. 
\Operation{Selective attention} focuses on relevant features while filtering noise. In the LEGO example, the child can attend to wing structure and stability while ignoring piece color or aesthetic details. This ability to focus resources and filter irrelevant input is fundamental to information processing, established by classic models demonstrating a bottleneck in early sensory processing that requires a selection mechanism \citep{broadbent1958perception, treisman1964monitoring}.
\Operation{Adaptive detail management} adjusts granularity based on task demands \citep{Rosch1978-ROSPOC-8}. Individuals dynamically shift between fine-grained and global representations based on cognitive goals, zooming in to examine how specific connection points distribute weight or zooming out to assess overall balance.
\Operation{Decomposition and integration} break problems into subproblems and synthesize solutions \citep{newell1959report}. Decomposition is a core mechanism of problem-solving, initially formalized through means-ends analysis and the creation of subgoals to manage task complexity. Integration represents the process of synthesizing findings across these subgoals to form a complete solution.
\Operation{Representational restructuring} reframes problems for new insight, fundamentally shifting how the problem is conceptualized and often marking breakthrough moments in reasoning \citep{Wertheimer1945, Kohler1925, braun2010structure}.
\Operation{Pattern recognition} detects recurring structures, enabling reuse of solution templates \citep{Selfridge1959, posner1968genesis}. This operation is the foundation of expert knowledge, enabling fast and accurate classification of novel stimuli by matching them against stored prototypes or templates. 
\Operation{Abstraction} generalizes from specific instances, distilling common, invariant features from a series of concrete examples and deriving principles from a set of actions or operations \citep{hull1920quantitative, piaget1952origins}.
After several construction attempts, the child may abstract the principle that heavier components require proportionally more connection points to remain stable.
Modification operations act on the representation of the problem, whereas the meta-cognitive controls of strategy selection and goal management (\S\ref{sec:taxonomy:controls}) act on the operations themselves, deciding which to deploy and toward which goal \citep{nelson1990metamemory}. Dividing the spaceship into body, wings, and cockpit changes the problem, while deciding to build the wings first changes the plan for solving it.

\textbf{Representation navigation} includes operations that traverse knowledge and inference structures.
\Operation{Forward chaining} reasons from known facts toward goals step-by-step 
\citep{huys2012bonsai}. This operation is a key component of early production systems and expert system architectures, where the system begins with facts and applies rules iteratively until a conclusion is reached or a goal state is satisfied \citep{newell1972human, davis1984origin}.
\Operation{Backward chaining} reasons from a goal back to known facts, repeatedly asking what must be true for the current goal to hold until it reaches conditions that are already satisfied 
\citep{park2017relative}. This goal-driven approach is efficient in situations where the number of possible outcomes is large but the goal is clearly defined \citep{charniak1985introduction, newell1959report}.
\Operation{Backtracking} revisits earlier reasoning upon detecting errors \citep{qin2025backtrack}. It often relies on depth-first search, where, upon hitting a dead-end, the system returns to the most recent decision point to explore alternatives \citep{Nilsson1971ProblemsolvingMI}.
In the LEGO example, the child uses forward chaining to build, backward chaining to plan, and backtracking when a design fails. These navigation operations determine the path through problem space.

\new{Table~\ref{tab:taxonomy} presents the complete taxonomy. Invariants specify the properties valid reasoning must have at the computational level, controls select and monitor strategies to achieve them at the resource-rational intermediate level, and representations and operations encode and transform knowledge at the algorithmic level. These 28 elements are foundational components extensively studied in cognitive science and give LLM researchers a vocabulary for examining how they manifest in human and LLM reasoning. %
Section~\ref{sec:main_experiment} uses this vocabulary to examine which elements appear in reasoning, \newtwo{how they connect into reasoning topologies}, and how they relate to success across problem types.}

\section{Behavioral Manifestation in Humans and LLMs}\label{sec:main_experiment}\vspace{-1.5mm}

\subsection{Methodology}\vspace{-1mm}

\subsubsection{Data Collection}
\label{sec:data_collection}

\begin{table}[h]
\centering\vspace{-2mm}
\caption{Overview of Data Sources and Models by Modality}\label{tab:datasets}\vspace{-3.5mm}
\scalebox{0.8}{\begin{tabular}{p{1.4cm} p{5.3cm} c p{6.7cm} c c}
\toprule
\textbf{Modality} & \textbf{Dataset} & \textbf{\#} & \textbf{Model} & \textbf{\#} & \textbf{\# Traces} \\
\midrule
\multirow{10}{*}{\textbf{Text}} & \multirow{9}{*}{GeneralThought \citep{RJT1990-GeneralThoughtArchive}} & \multirow{9}{*}{10{,}307} & Qwen3 (8B, 14B, 32B) \citep{qwen3} & 3 & 31,791 \\
& \multirow{9}{*}{ClaimSpect \citep{kargupta-etal-2025-beyond}} & \multirow{9}{*}{290} & DeepSeek-R1-Distill-Qwen2.5 (1.5B, 7B, 14B, 32B) \citep{deepseek_r1} & 4 & 42,388 \\
& & & DeepSeek-R1-Distill-Llama3 (8B, 70B) \citep{deepseek_r1} & 2 & 21,194 \\
& & & DeepScaleR-1.5B-Preview \citep{deepscaler2025} & 1 & 10,597 \\
& & & s1.1-32B \citep{muennighoff2025s1simpletesttimescaling} & 1 & 10,597 \\
& & & OpenThinker-32B \citep{guha2025openthoughtsdatarecipesreasoning} & 1 & 10,597 \\
& & & DeepHermes-3-Llama-3-8B-Preview \citep{hermes1} & 1 & 10,597 \\
& & & DeepSeek-R1 \citep{deepseek_r1} & 1 & 10,597 \\
& & & Olmo~3 (7B, 32B) \citep{olmo3} & 2 & 21,194 \\
& \multicolumn{1}{l}{\textbf{Total (Text Data)}} & \textbf{10{,}597} & \multicolumn{1}{l}{\textbf{Total (Text Models)}} & \textbf{16} & \textbf{169,552} \\
\midrule
\multirow{4}{*}{\textbf{Audio}} & BLAB \citep{ahia2025blabbrutallylongaudio} & \new{386} & Qwen3-Omni-30B-A3B-Thinking & 1 & 4,660 \\
& MMAR \citep{ma2025mmar} & \new{870} & \cite{qwen3} & & \\
& MMAU-Pro \citep{kumar2025mmaupro} & \new{3{,}404} & & & \\
& \multicolumn{1}{l}{\textbf{Total (Audio Data)}} & \textbf{4{,}660} & \multicolumn{1}{l}{\textbf{Total (Audio Models)}} & \textbf{1} & \textbf{4,660} \\
\midrule
\multirow{2}{*}{\textbf{Image}} & \texttt{Zebra-CoT} \citep{li2025zebracot} & 18{,}000 & \new{Curated traces refined with GPT-4.1 and Gemini-2.5-Pro} & 1 & 18,000 \\
& \multicolumn{1}{l}{\textbf{Total (Image Data)}} & \textbf{18{,}000} & \multicolumn{1}{l}{\textbf{Total (Image Models)}} & \textbf{1} & \textbf{18,000} \\
\midrule
\multicolumn{2}{l}{\textbf{Grand Total (All Data)}} & \textbf{33{,}257} & \multicolumn{1}{l}{\textbf{Grand Total (All Models)}} & \textbf{18} & \textbf{192,212} \\
\midrule
\textbf{Human} & Story, diagnosis-solution, & \new{30} & Human participants & \new{--} & \new{1,346} \\
\textbf{Matched} & and case analysis problems & & Text models above, excluding DeepSeek-R1 & \new{15} & \new{450} \\
\bottomrule
\end{tabular}\vspace{-2mm}
}
\end{table}

\textbf{LLM reasoning data.}
\new{For text, we draw problems from \texttt{GeneralThought} \citep{RJT1990-GeneralThoughtArchive}, a large collection of reasoning problems across many tasks and domains, and down-sample its most common tasks. \texttt{GeneralThought} consists mostly of problems with verifiable answers and contains almost no dilemmas,\footnote{Dilemmas require resolving situations with contradictory positions and no clear satisfactory solution (Appendix~\ref{app:problem_types}).} so we add 290 biomedical and geopolitical claims from \texttt{ClaimSpect} \citep{kargupta-etal-2025-beyond}, rewritten as questions. Most of these are dilemmas, and the rest are mainly design and case analysis problems. Each of the resulting 10,597 problems is solved by 16 reasoning models from several model families and training recipes (Table~\ref{tab:datasets}), so that our findings do not depend on any one of them. 
We re-use the \texttt{DeepSeek-R1} traces released with \texttt{GeneralThought} and generate \texttt{DeepSeek-R1} traces only for the \texttt{ClaimSpect} problems.}

\new{For audio, we use problems from BLAB, which tests reasoning over long recordings \citep{ahia2025blabbrutallylongaudio}, MMAR, which covers a range of audio tasks \citep{ma2025mmar}, and MMAU-Pro, which covers a range of listening skills \citep{kumar2025mmaupro}. They are solved by \texttt{Qwen3-Omni-30B}, the only open-weight audio-language model with a thinking mode, and we keep the 4,660 problems on which it produced a reasoning trace (386 from BLAB, 870 from MMAR, and 3,404 from MMAU-Pro).}

\new{For images, we use \texttt{Zebra-CoT} \citep{li2025zebracot}, a dataset of image problems paired with curated reasoning traces, and sample 1,000 problems from each of its 18 task types. \hypertarget{zebra_cot_desc}{Some of its traces come from existing math, physics, coding, and chess datasets, and the rest were written from templates and refined by frontier models (\texttt{Gemini-2.5} and \texttt{GPT-4.1}).} These 18,000 traces also allow us to analyze which cognitive elements appear in synthetic training data.}
\new{Table~\ref{tab:datasets} summarizes the problems, models, and traces in each modality, 192,212 traces in total.}

\textbf{\new{Matched human-model data.}}
\new{To compare human and model reasoning on the same problems, we curate 30 problems for this study, ten story problems (well-structured), ten diagnosis-solution problems (\newtwo{intermediate}), and ten case analysis problems (ill-structured). We balanced the problems for difficulty and wrote them so that solving them requires no domain expertise. Participants recruited on Prolific under an IRB-approved protocol solved randomly assigned problems while thinking aloud, and their recordings were transcribed automatically \citep{evernote_ai_nodate}, obtaining \newthree{1,346} human traces.
For the same 30 problems, we sampled one trace from each of the 15 text models other than \texttt{DeepSeek-R1}, giving 450 model traces. Human data collection details are in Appendix~\ref{app:prolific}.
}

\subsubsection{Fine-Grained Cognitive Element Annotation}
\new{\label{sec:annotation}}

We annotate each reasoning trace using the 28-cognitive-element taxonomy (Section~\ref{sec:taxonomy}). %
For each cognitive element, we develop annotation guidelines that include: 1) operational definitions grounded in the cognitive science literature, 2) \new{concrete indicators of how the element appears in the text}, 3) three-level scoring rubrics, 4) manually curated in-context examples with explanations, and 5) span identification requirements using character indices.
\new{This protocol is developed iteratively with human-in-the-loop feedback on a pilot set of human reasoning traces (Appendix~\ref{app:human_pilot}).} 
We use \texttt{GPT-4.1} to produce the span annotations with the finalized guidelines. 
For each trace and each of the 28 elements, the annotation returns the spans that exhibit the element and a \textit{presence score} on the rubric: absent (0), partially present (1), or present (2).

\subsubsection{Problem Type Classification \& Response Evaluation}
\new{\label{sec:problem_eval}}

\textbf{Jonassen's Problem Taxonomy.}
We extend the problem-solving taxonomy in \citet{jonassen2000toward} to classify cognitive tasks in our dataset. 
\citet{jonassen2000toward} defines problem-solving as a goal-directed cognitive activity that transforms an \textit{initial state} into a \textit{desired goal state} through systematic reasoning, proposing a typology of 11 problem types: logical problems, algorithms, story problems, rule-using, decision-making, troubleshooting, diagnosis-solution, strategic performance, case analysis, design, and dilemmas. 
\citet{jonassen2015all} further characterizes ten out of the eleven problem types along a continuum from well-structured (clear goals, known solution paths, predictable outcomes) to ill-structured (ambiguous goals, multiple solution paths, uncertain outcomes). 
However, some tasks in the \texttt{GeneralThought} dataset do not require goal-directed transformation. 
We add two categories to capture tasks outside this paradigm: factual recall (retrieving stored knowledge without reasoning) and creative/expressive tasks (generating novel content judged by originality or aesthetic quality rather than convergence to a predetermined solution). 
This yields a \textbf{13-category typology} (Appendix~\ref{app:problem_types}). 
Note that we drop the strategic performance problem type from \citet{jonassen2000toward}, which requires multi-turn interaction with an environment.
Problem type labels are assigned to each task in our dataset using \texttt{o4-mini} (high reasoning effort).

\textbf{Response Correctness.}
AlpacaEval \citep{dubois2024length} with \newthree{\texttt{GPT-4.1}} as the judge is used to assess response correctness. For each problem-response pair, the LLM judge receives the original problem, the model's response, and a reference response. Ground truth answer is used as the reference for verifiable tasks and \texttt{Claude-Sonnet-4.5}'s response (selected for strong benchmark performance) for non-verifiable tasks.

\subsubsection{\new{Metrics}}
\new{\label{sec:measures}
All metrics are computed from the presence scores produced by the span annotation (Section~\ref{sec:annotation}):
\begin{itemize}[itemsep=0pt,topsep=-2pt,leftmargin=16pt]
    \item \new{\textit{Presence}: how often an element appears in a set of traces. An element is \textit{clearly present} in a trace if its presence score is 2. We report (a) \newthree{the \textit{length-normalized presence rate}}, the fraction of traces in which an element is clearly present, and (b) the \new{\textit{length-normalized mean presence score}}, which assigns partial score to partial presence annotations. \newthree{Both divide each trace's contribution by the logarithm of the trace's length, because longer traces exhibit more elements (Appendix~\ref{app:length}).}
    }
    \item \textit{Allocation}: an element's presence score divided by the sum of presence scores over all 28 elements in the trace, \new{which describes how a trace divides its reasoning among elements regardless of its length. We measure how much two allocation profiles differ by half their L1 distance, which is the fraction of reasoning that the two profiles assign to different elements.}
    \item \textit{Association with success}: \new{positive pointwise mutual information (PPMI) between an element's presence and success for each problem type, normalized PMI (NPMI) for reasoning \newtwo{topologies} (Section~\ref{sec:pattern_extraction}). %
    }
\end{itemize}
}

\subsubsection{Reasoning \newtwo{Topology} Construction}
\label{sec:pattern_extraction}

\par \new{We represent each trace as a graph of its cognitive elements to visualize the reasoning process. Each node is an element in the trace, and each edge records whether one element's text span follows the other's (\texttt{NEXT}), lies inside it (\texttt{CONTAINS}), or covers nearly the same text (\texttt{PARALLEL}).}
Concretely, for two spans \newthree{$b_a = [s_a, e_a]$} and \newthree{$b_b = [s_b, e_b]$} with $s_a \leq s_b$, we set $\tau_{par} = 20$ chars and $\tau_{overlap} = 0.8$ and assign the edge as \newthree{follows}.
\begin{enumerate}[itemsep=0pt,topsep=-2pt,leftmargin=16pt]
    \item \texttt{PARALLEL} (the spans cover nearly the same text): if $|s_b - s_a| + |e_b - e_a| \leq \tau_{par}$, or if $b_a$ extends past $b_b$ ($e_a > e_b$) and $b_b$ covers at least a fraction $\tau_{overlap}$ of it. 
    \item \texttt{CONTAINS} \new{(most of $b_b$ lies inside $b_a$): if $s_b \leq e_a \leq e_b$ and $\frac{e_a - s_b}{e_b - s_b} \geq \tau_{overlap}$.}
    \item \texttt{NEXT} \new{($b_b$ follows $b_a$): in all other cases\newtwo{, keeping only the first such $b_b$ after $b_a$, so that each span links to the span that directly follows it}.}
\end{enumerate}

\textbf{Reasoning topology.} To compare reasoning across many traces, we summarize the graphs of a set of traces (for example, all model traces on one problem type) into a single graph of seven elements, which we call a \textbf{\emph{reasoning topology}}. Starting from one element, we repeatedly follow an outgoing edge to an element not yet in the topology, until the topology has seven elements or no such edge remains. 
Specifically, we construct \textbf{(1) typical topology ($\hat{G}$)} starting from the element whose span most often comes first, and each step takes the edge that appears most frequently; and \textbf{(2) successful topology ($G^*$)} starting from the element most associated with success (the highest NPMI) among those that begin correct traces, and each step takes the edge with the highest NPMI, and stops early when every remaining edge has non-positive NPMI.

\subsection{Experimental Setup}

\subsubsection{Dataset Composition}

\begin{figure}[t]
     \centering
     \begin{subfigure}[b]{0.52\textwidth}
         \centering
         \new{\includegraphics[width=\textwidth]{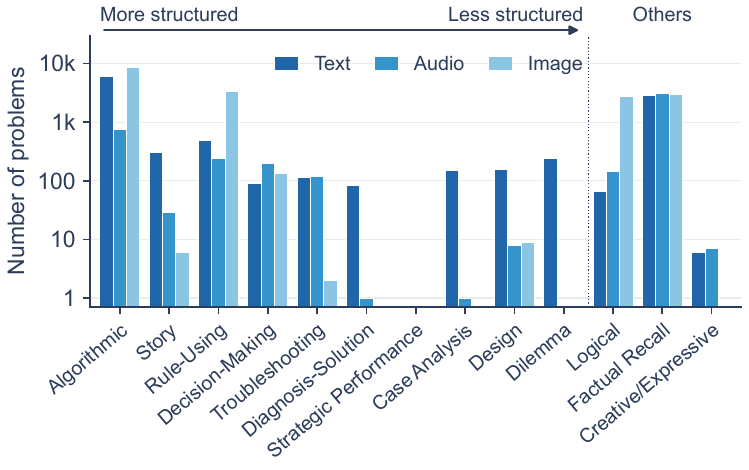}}\vspace{-2mm}
         \caption{\small Problem Type Distribution by Modality}
         \label{fig:problem_type}
     \end{subfigure}
     \hfill
     \begin{subfigure}[b]{0.46\textwidth}
         \centering
         \new{\includegraphics[width=\textwidth]{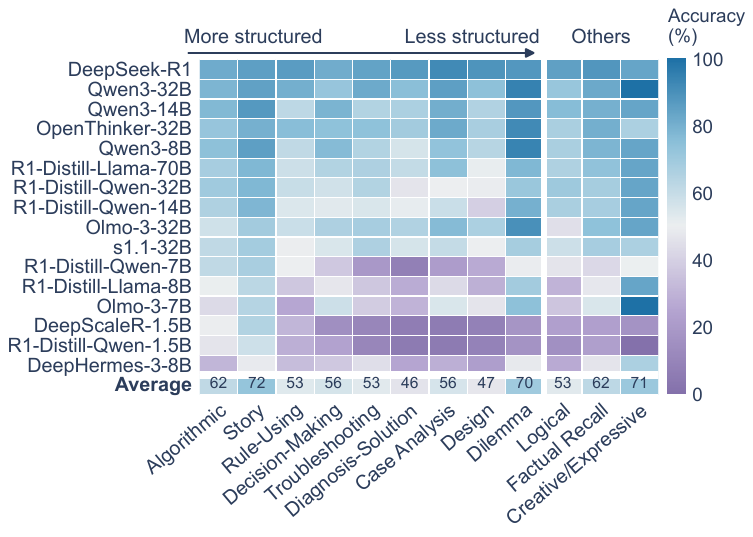}}\vspace{-2mm}
         \caption{\small Accuracy by Problem Type and Model}
         \label{fig:problem_difficulty}
     \end{subfigure}
     \vspace{-2.5mm}
     \caption{\small{Dataset composition and difficulty across problem types, ordered along Jonassen's structuredness continuum. 
     Accuracy is lower and varies more across models on less structured problems. Details in Appendix~\ref{app:accuracy_analysis}.
     }}\vspace{-2mm}
     \label{fig:data_stats}
\end{figure}

\textbf{Problem type.}
Figure~\ref{fig:problem_type} shows the number of problems of each type in each modality. 
Algorithmic and factual recall problems are the most common, making up 47\% and 27\% of all problems.
Ill-structured problems come almost entirely from text (\newtwo{539} text problems, against 9 image and \newtwo{9} audio problems), so our analyses of problem structure rest on the text models. 

\textbf{Problem difficulty.}
Success rates vary by problem structure (Figure~\ref{fig:problem_difficulty}). Averaged accuracy is 62.4\% on well-structured problems, 51.7\% on intermediate problems, and 57.5\% on ill-structured problems. 
Accuracy across models differ more on intermediate and ill-structured problems than well-structured problems. 

\subsubsection{\new{Research Questions}}\vspace{-2mm}
\new{\label{sec:rqs}}

\par \new{We organize the analysis around the following research questions. We answer each on the large model corpus and use the matched corpus to compare model reasoning with human reasoning.}
\begin{enumerate}[itemsep=0pt,topsep=-3pt,leftmargin=16pt]
    \item \new{\textbf{\textit{What does reasoning consist of?}} Which elements appear in reasoning traces, regardless of outcome, and how this varies across problem types, across models, and between models and humans.}
    \item \new{\textbf{\textit{What makes reasoning succeed?}} Which elements distinguish correct from incorrect traces, and whether correct and incorrect reasoning differ in the same way for humans and models.}
    \item \new{\textbf{\textit{How does reasoning change with problem structure?}} Which elements \newthree{predict success} as problems lose structure, and whether human and model reasoning change accordingly.}
    \item \new{\textbf{\textit{\newtwo{What is the topology of reasoning?}}} How elements are ordered within a trace, and whether the \newthree{typical} order matches the order of successful traces.}
\end{enumerate}

\subsection{Results \& Analyses}\vspace{-2mm}\label{sec:results}

\begin{figure}[t]
\begin{minipage}{\textwidth}
\begin{minipage}{0.52\textwidth}
    \centering
    \newthree{\includegraphics[width=\linewidth,height=0.8\linewidth,keepaspectratio]{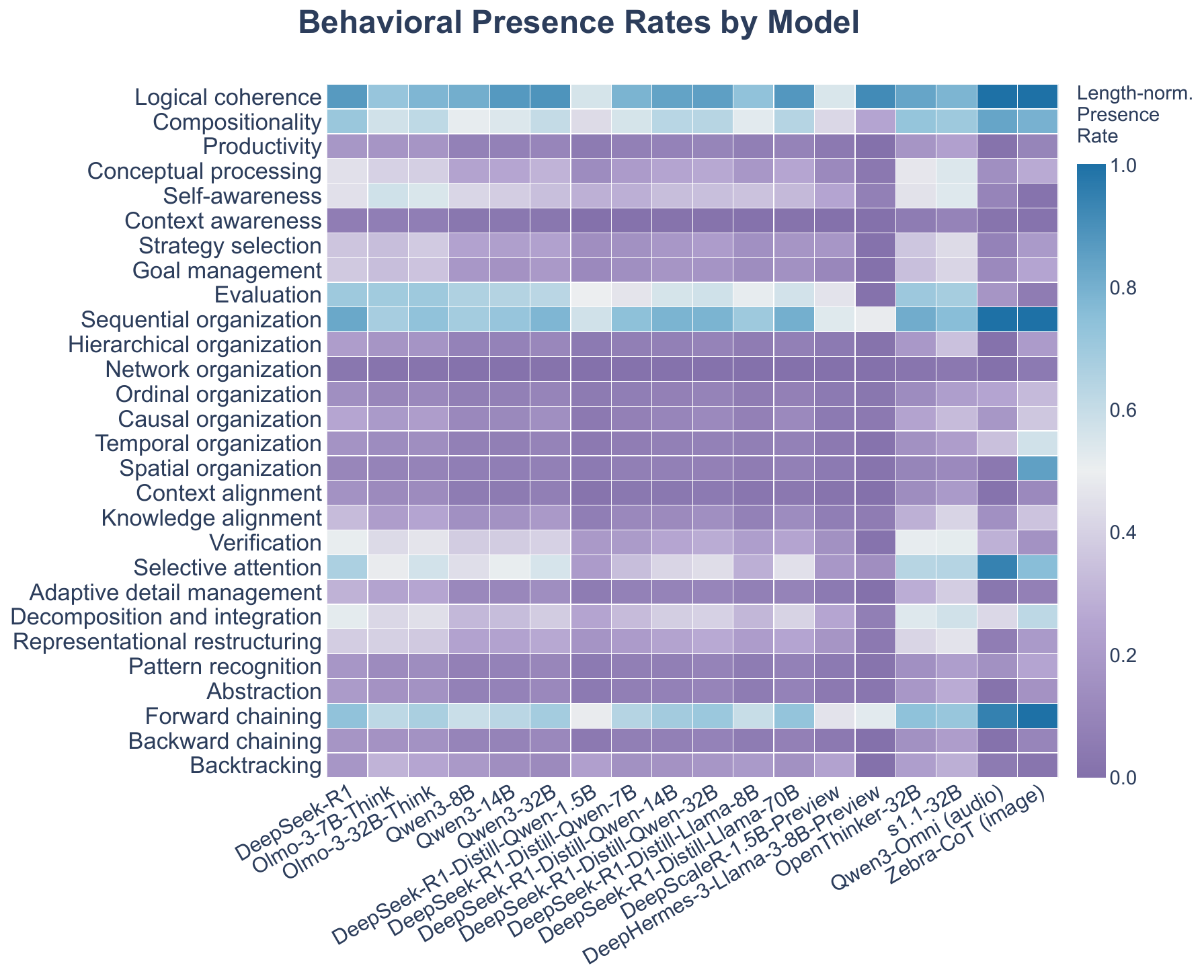}}
    \end{minipage}
    \begin{minipage}{0.48\textwidth}
    \centering
    \vspace{4mm}
    \new{\includegraphics[width=\linewidth,height=0.8\linewidth,keepaspectratio]{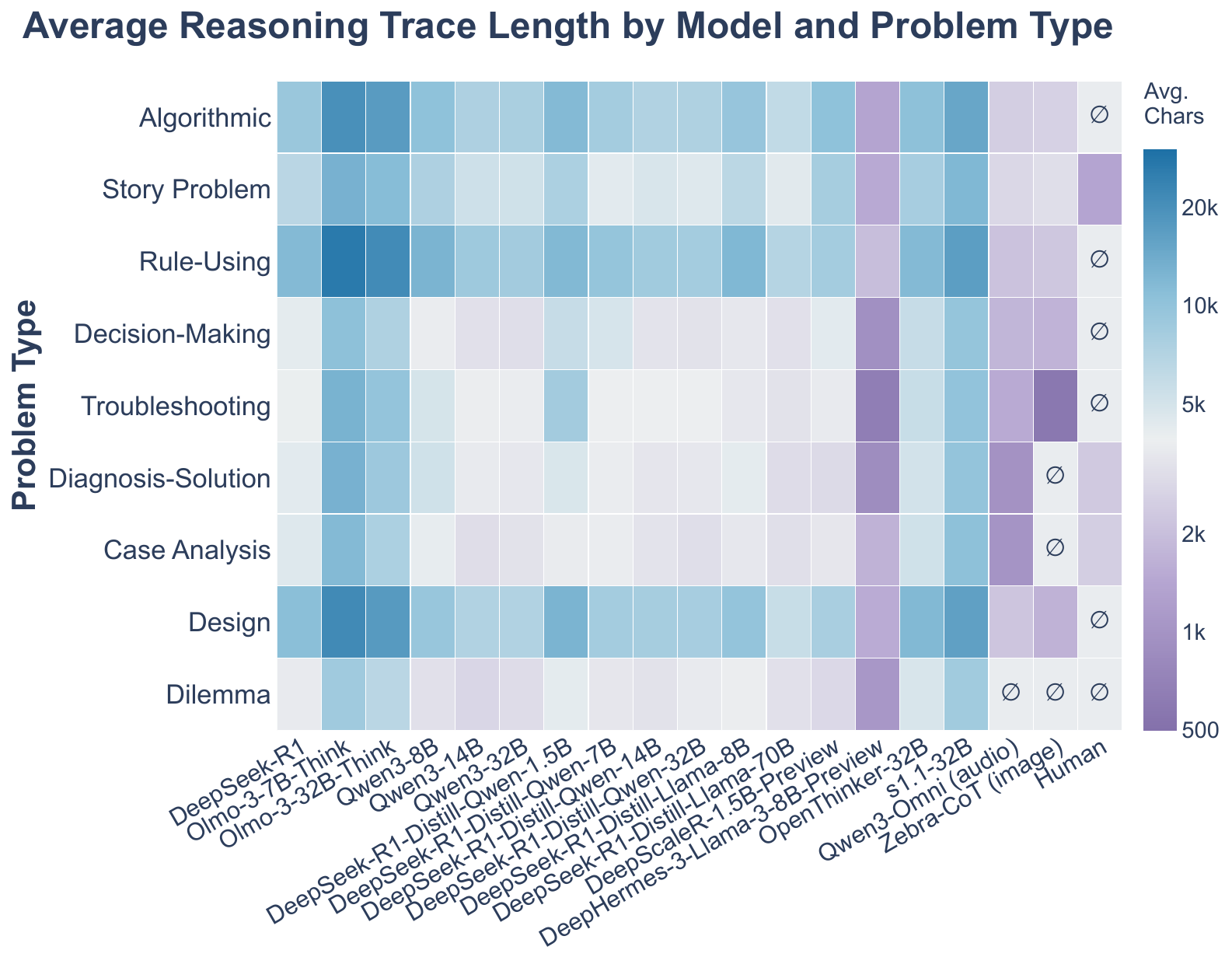}}
    \label{fig:trace_length}
\end{minipage}
\end{minipage}\vspace{-5mm}
\caption{\small
\textbf{\textit{(Left)}} Length-normalized presence rate of each cognitive element in the traces of each model. \textbf{\textit{(Right)}} Average reasoning trace length (\# characters, log colour scale) for each model and problem type; cells marked $\emptyset$ have no traces. All models share a similar element profile and differ mainly in how many elements they exhibit.}
\label{fig:model_behavior_presence}\vspace{-3mm}
\end{figure}

\subsubsection{Distribution of Cognitive Elements in Reasoning Traces}\vspace{-1mm}
\label{sec:behavior_distribution}

\textbf{\new{Different models exhibit cognitive elements similarly.}} 
Models differ mainly in the density of cognitive elements in its reasoning traces (Figure~\ref{fig:model_behavior_presence}, left), and the verbosity of their reasoning (Figure~\ref{fig:model_behavior_presence}, right). 
\newthree{All text models meaningfully use at least 25 of the 28 elements (length-normalized presence rate of 0.05 or more), except \texttt{DeepScaleR-1.5B} (15), \texttt{R1-Distill-Qwen-1.5B} (16), and \texttt{DeepHermes-3}, the least accurate model overall, which exhibits only 8.} 
Frequency rankings of the elements are almost identical for each model (\newthree{$\rho=.92$}). 
\Invariant{Logical coherence}, \Representation{sequential organization}, and \Operation{forward chaining} are among the most frequent elements in every model, and \Metacognitive{context awareness} is among the rarest. 
The audio and image traces fall within the range of the text models and share their three most frequent elements, and the image traces also show frequent \Representation{spatial organization}.

\begin{wrapfigure}{R}{0.497\textwidth}
    \centering
    \vspace{-3mm}
    \includegraphics[width=\linewidth]{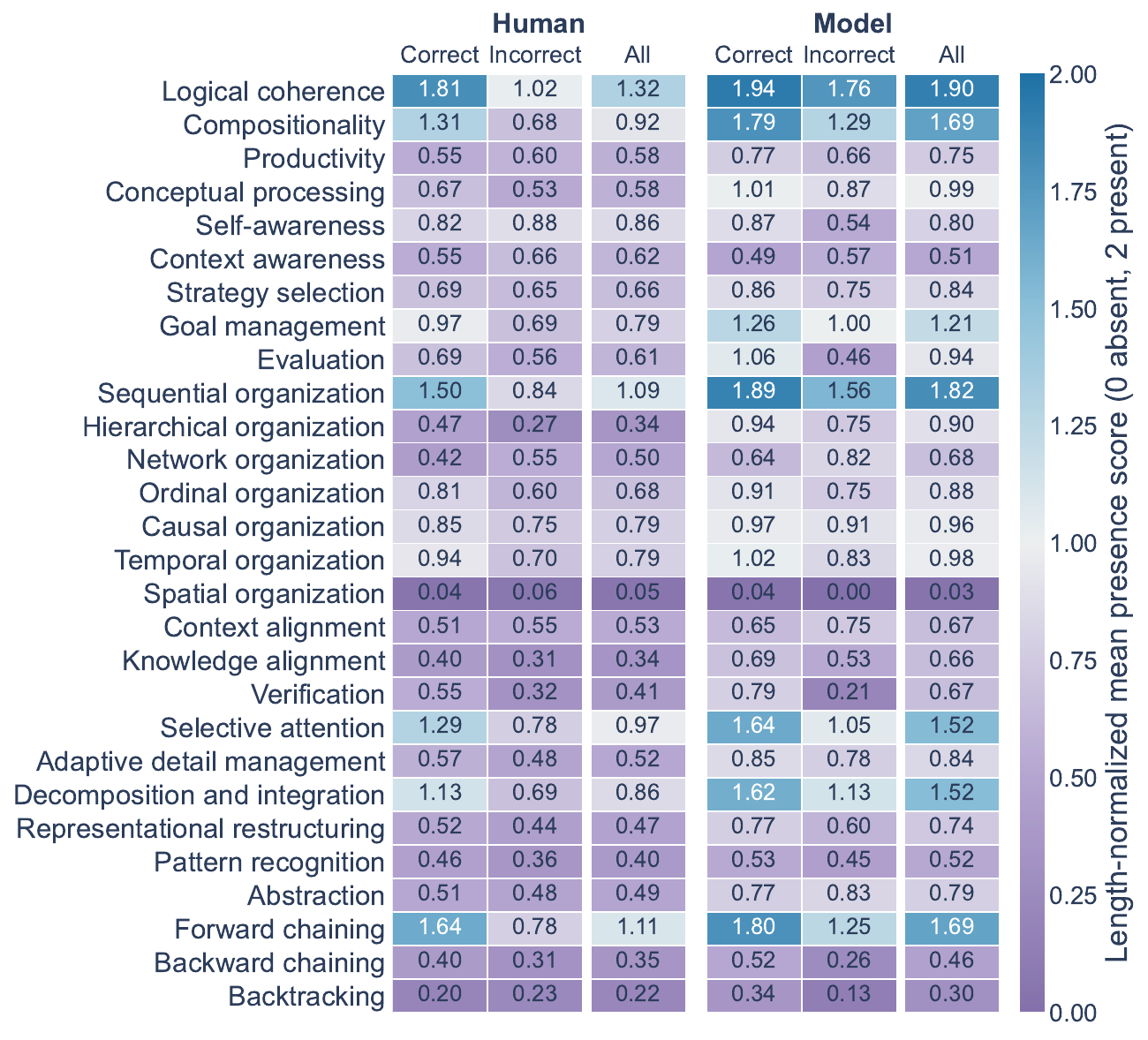}\vspace{-4.2mm}
    \caption{\small{Length normalized mean presence score. 
    Most elements appear more often in model traces. 
    Correct traces contain denser elements than incorrect ones in both models and humans ($p<10^{-17}$).
    }}
    \label{fig:prolific_heatmap}\vspace{-5mm}
\end{wrapfigure}

\textbf{Models match humans in element coverage but not distribution.} 
Model reasoning is similar to that of humans in the \emph{coverage} of cognitive elements. 
On the matched corpus, every cognitive element used by humans also appears in model traces.
However, we find that model reasoning differs from human reasoning in the density and frequency of the cognitive elements.
First, cognitive elements are more densely packed in model reasoning than human reasoning. 
After normalizing by log trace length, 25 of the 28 elements appear more often in model traces (Figure~\ref{fig:prolific_heatmap}). 
Second, the two elements that appear more often in human traces are both meta-cognitive controls---\Metacognitive{context awareness} (21.6\%$\uparrow$), which is among the rarest elements in every model, and \Metacognitive{self-awareness} (7.5\%$\uparrow$)---whereas models extensively use \Representation{sequential organization} (67.0\%$\uparrow$). 
Moreover, spoken reasoning leaves well-practiced steps unsaid \citep{ericsson1980verbal}, so human traces could understate parts of the human cognitive process that happen subconsciously. This could partially explain the higher presence of most elements in model traces, and makes the two meta-cognitive exceptions more notable. The differences between human and model reasoning therefore lie in \textit{\textbf{\newthree{meta-reasoning}}}, \newthree{the monitoring and control of one's own reasoning, including when elements are deployed and how their outcomes are checked}, which the rest of this section examines. Table~\ref{tab:prolific_traces} shows a qualitative example of human and model reasoning traces.

\begin{finding}
\new{Models exhibit nearly every cognitive element seen in human reasoning, but differ in how the elements are used.}
\end{finding}

\begin{table}[t]
\captionof{table}{\small Example diagnosis-solution problem. \newtwo{Both human and model reach the same diagnosis with shared elements, but the model packs in more elements and proceeds step by step as an outside observer, while the human reasons from \newthree{their} own feelings, with the context awareness and self-awareness that are more frequent in human traces.}
}
\label{tab:prolific_traces}\vspace{-2mm}
\small
\centering
\renewcommand{\tabularxcolumn}[1]{m{#1}} %
\resizebox{\linewidth}{!}{%
\begin{tabularx}{1.23\textwidth}{|l|>{\raggedright\arraybackslash}X|}
\hline
\makecell[l]{\textbf{Problem}\\\textbf{summary}} & \newtwo{\textit{You feel disconnected from a close friend: they take 24--48 hours to answer your texts, yet are engaged in person, active on social media, and say they are ``bad at texting.'' Your other friends reply within hours, and you have started to feel hurt. Diagnose what's happening and propose how to address this.}} \\
\hline
\textbf{Reasoner} & \textbf{Reasoning Trace Excerpt} \\
\hline
\makecell[l]{Human \\ (285 words)} & 
\Operation{so what i think is happening is that there's a mismatch in communication styles not the lack of care} \Operation{\textbf{<selective attention>}} my friend is really present and engaged when we are together in person ... \Metacognitive{at the same time} 
\Metacognitive{they are slow texting especially when i see them acting on social media makes it feel personal even if} 
\Metacognitive{they say it isn't \textbf{<context awareness>}} \Metacognitive{I think the reason it hurts is because timely texting is important} 
\Metacognitive{to me. It's how I stay connected and reassured. ... Even though logically, I know that might not be} 
\Metacognitive{true. \textbf{<self-awareness>}} The way I address it is by being honest but calm. ... \Metacognitive{If after communication, clearly,} 
\Metacognitive{I still feel hurt over time, then I'll have to reflect on whether this friendship is meeting my emotional} 
\Metacognitive{needs. \textbf{<evaluation>}} ... \\
\hline
\newtwo{\makecell[l]{Olmo-3-7B \\ (744 words)}} & 
\Metacognitive{Okay, so I need to figure out what's going on with this friend and how to address the disconnection.} 
\Metacognitive{\textbf{<goal management>}} \Operation{Let's start by breaking down the situation the user described. First, the friend}
\Operation{takes 24-48 hours to respond to texts, sometimes even longer. \textbf{<decomposition>}} ... \Operation{The key here is the} 
\Operation{discrepancy between in-person interaction and text communication. \textbf{<selective attention>}} ... \Operation{Possible} 
\Operation{diagnoses: The issue is likely a mismatch in communication styles between the two friends.} 
\Operation{\textbf{<pattern recognition>}} ... \Representation{Possible steps: 1. Have a calm, non-confrontational talk with the friend.}
\Representation{2. Express feelings honestly but kindly. 3. ... \textbf{<sequential organization>}} ... \\
\hline
\end{tabularx}
}\vspace{-1mm}
\end{table}

\subsubsection{Cognitive Elements Predict Correctness Better in Humans Than in Models}\vspace{-1mm}
\new{\label{sec:success}}

\textbf{Humans and models share success predictors.}
Answer correctness is correlated with element density and certain cognitive elements in both humans and models.
Correct traces contain more elements than incorrect ones after length normalization (Figure~\ref{fig:prolific_heatmap}).
The elements most predictive of success are also largely the same in both groups, led by \Operation{forward chaining}, \Invariant{compositionality}, \Operation{selective attention}, and \Operation{decomposition}.
Because humans and models succeed with the same elements despite reasoning in very different ways, these elements could reflect what the problems demand of any reasoner. The taxonomy thus captures properties of correct reasoning that hold across reasoners beyond human reasoning.

\begin{wrapfigure}{r}{0.37\textwidth}\vspace{-3mm}
    \centering
    \new{\includegraphics[width=\linewidth]{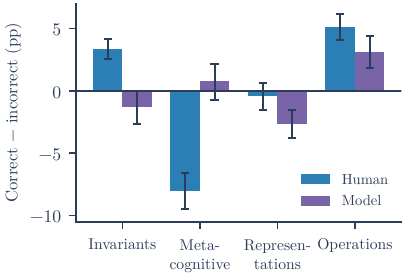}}\vspace{-3mm}
    \caption{\small{Difference in allocation between correct and incorrect traces is larger in humans than in models.}}
    \label{fig:allocation_correctness}\vspace{-4mm}
\end{wrapfigure}

\label{sec:regulation}\textbf{Correct and incorrect traces differ more in humans.}
Grouped by correctness, human reasoning traces differ significantly for most \newthree{dimensions}, and the difference is greater in humans than in \newthree{models} (Figure~\ref{fig:allocation_correctness}). 
Humans especially allocate more of their reasoning to \Metacognitive{meta-cognitive controls} in traces that lead to incorrect answers (25\%) compared to correct answers (17\%). We further find that this difference comes mainly from an increase in \Metacognitive{self-awareness} in incorrect traces, where human participants sometimes say \textit{``I am not good at this type of problems.''} in their reasoning.
In models, \Invariant{invariants} and \Metacognitive{meta-cognitive controls} show no significant difference. %
At the element level, leading up to incorrect answers, models \newthree{tend} to decrease their effort in \emph{self-monitoring}---\Operation{verification} of individual steps drops by 71\% and \Metacognitive{evaluation} of overall progress drops by 44\%---while increasing their effort in \Representation{network organization}, \Operation{abstraction}, and \Operation{context alignment}.
In other words, incorrect model traces spend relatively more of their reasoning relating concepts and the problem context and less on \newthree{checking it}.

Notably, \Invariant{logical coherence} distinguishes correct from incorrect traces far more in humans (Figure~\ref{fig:prolific_heatmap}).
Its length-normalized presence is 77\% higher in correct human traces than in incorrect ones, while in model traces it is present almost regardless of outcome and differs by only 0.18, which is 22.8\% of the human gap.
\newthree{In humans, the trace follows the cognitive process, so reasoning that fails also loses coherence.}
\newthree{Language models could instead produce locally consistent text by default, so a model trace can remain coherent when the reasoning behind it fails, which hides the failure from anyone reading the trace and limits how well model reasoning can be monitored \citep{guan2025monitoringmonitorability}.}

\begin{finding}
The same cognitive elements predict success in humans and models, but human traces reveal failure more clearly (less coherent and more self-aware).
\end{finding}

\begin{figure}[t]
\new{\begin{minipage}{0.5\textwidth}
    \centering
    \newthree{\includegraphics[width=0.95\linewidth]{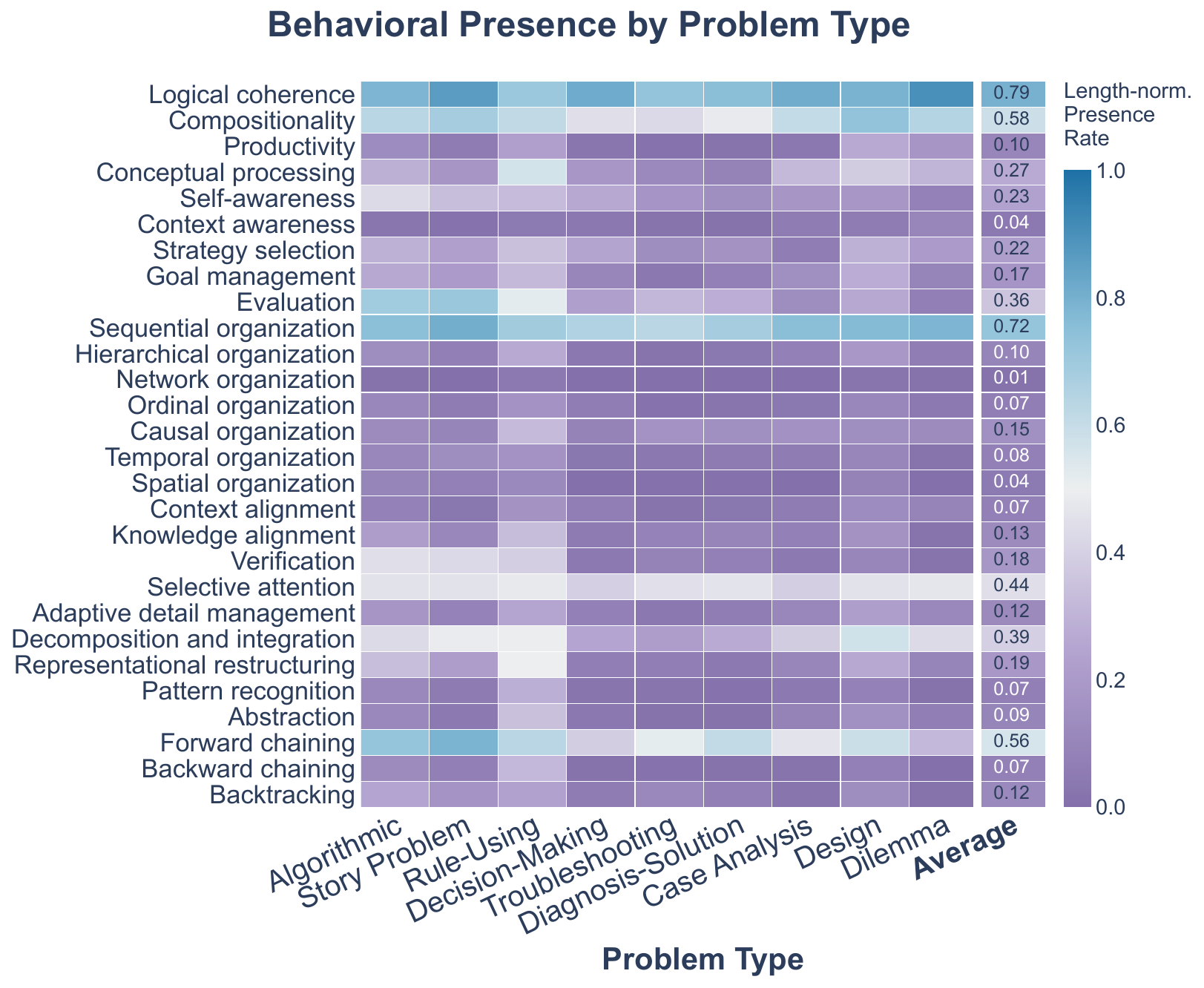}}
\end{minipage}%
\hfill
\begin{minipage}{0.5\textwidth}
    \centering
    \includegraphics[width=0.93\linewidth]{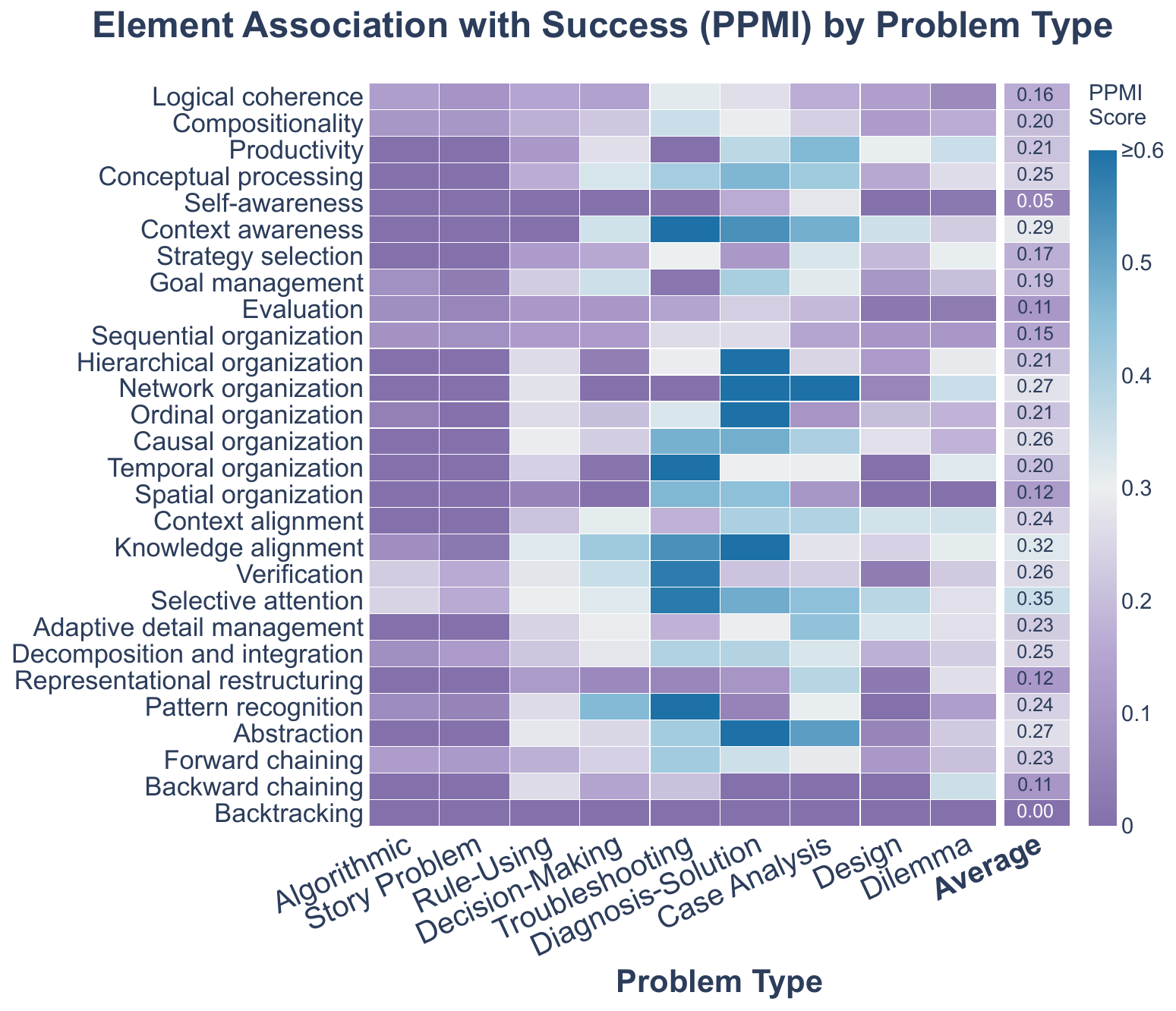}
\end{minipage}}\vspace{-3mm}
\caption{\small
Problem type analysis. \textbf{\textit{(Left)}} \newthree{Length-normalized presence rate of each element} for each problem type, ordered from well-structured to ill-structured. \textbf{\textit{(Right)}} Positive pointwise mutual information (PPMI) between the presence of each element and trace success. %
Elements predict success more on less structured problems.}\label{fig:presence_ppmi_heatmaps_v5}
\end{figure}

\begin{figure}[t]
    \centering
    \includegraphics[width=\linewidth]{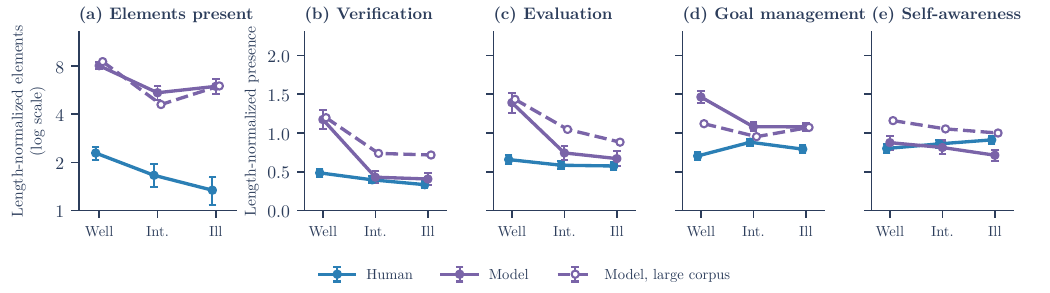}\vspace{-4mm}
    \caption{\small
    \textbf{\textit{(a)}} Length-normalized element presence per trace. 
    Both humans and models use fewer elements on less structured problems. \textbf{\textit{(b--e)}} Length-normalized mean presence score of \Operation{verification}, \Metacognitive{evaluation}, \Metacognitive{goal management}, and \Metacognitive{self-awareness}. Models monitor their own reasoning less on less structured problems, while \newthree{humans} consistently self-monitor. Per-element details in App.~\ref{app:structure_by_element}.
    }
    \label{fig:prolific_adaptation_regulation}\vspace{-2mm}
\end{figure}

\subsubsection{Reasoning Differences Across Problem Types}\vspace{-1mm}
\new{\label{sec:problem_type}}

\textbf{Presence of cognitive elements is more predictive of success in \newtwo{less structured} problems.}
Figure~\ref{fig:presence_ppmi_heatmaps_v5} (right) reports the association between the presence of elements and the correctness of the \newthree{trace}.
\newthree{Averaged over all cognitive elements, the association} with success is \newtwo{2.4} times stronger in the \newtwo{three} ill-structured problem types (PPMI=\newtwo{0.217}) than in the well-structured algorithmic, story, and rule-using problems (PPMI=0.091)\newtwo{, and strongest in the intermediate types (PPMI=0.298)}.
This suggests that knowing the right cognitive elements to use is crucial for success on \newtwo{less structured} problems, \newthree{whereas} the solution procedure is more defined and known on well-structured problems, so success could depend more on executing it accurately, 
which the presence of an element does not capture.

\par \textbf{Humans and models both use fewer elements in ill-structured problems, but differ in which elements they keep.}
As problems become less structured, the average presence of cognitive elements drops in both the large corpus (Figure~\ref{fig:presence_ppmi_heatmaps_v5}) and the matched corpus (Figure~\ref{fig:prolific_adaptation_regulation}a).
Human traces contain 41\% fewer elements in ill-structured problems than in well-structured ones, and model traces 26\% fewer, with both groups using less \textit{step-by-step reasoning} (\Operation{forward chaining}, \Representation{sequential organization}).

However, we find that the two groups differ in \newthree{meta-reasoning}.
First, human reasoning in ill-structured problems contains more \Metacognitive{goal management} (+0.09) and \Metacognitive{self-awareness} (+0.11), while model reasoning contains less of both ($-0.39$ and $-0.16$; Figure~\ref{fig:prolific_adaptation_regulation}d,e).
\textbf{\textit{Human \newthree{meta-reasoning} thus increases with uncertainty}}, as reasoners track their goals and their own limits more closely when the problem offers no fixed procedure or a way to check the answer \citep{flavell1979metacognition}.
Second, models monitor their reasoning (\Operation{verification} and \Metacognitive{evaluation}) much less in ill-structured problems than in well-structured ones, while humans monitor theirs at similar levels (Figure~\ref{fig:prolific_adaptation_regulation}b,c).
\textbf{\textit{Model \newthree{meta-reasoning} instead follows verifiability}}, as models monitor their reasoning less where an answer cannot be recomputed, consistent with self-verification acquired through training on verifiable rewards \citep{gandhi2025cognitive}.

\begin{finding}
Humans and models both use fewer elements on ill-structured problems, 
but human \newthree{meta-reasoning} increases with uncertainty while model \newthree{meta-reasoning} follows verifiability.
\end{finding}

\subsubsection{Reasoning \newtwo{Topologies}: How Cognitive Elements Interact}
\label{sec:behavior_structures}

Beyond which elements appear, reasoning is characterized by the order in which elements follow each other within a trace. 
Thus, we extract transition graphs between elements (\S\ref{sec:pattern_extraction}) and compare the typical \newtwo{topology} in model traces ($\hat{G}$) with the \newtwo{topology} most associated with success ($G^*$).

\begin{figure}[t]
    \centering
    \includegraphics[width=1.0\linewidth]{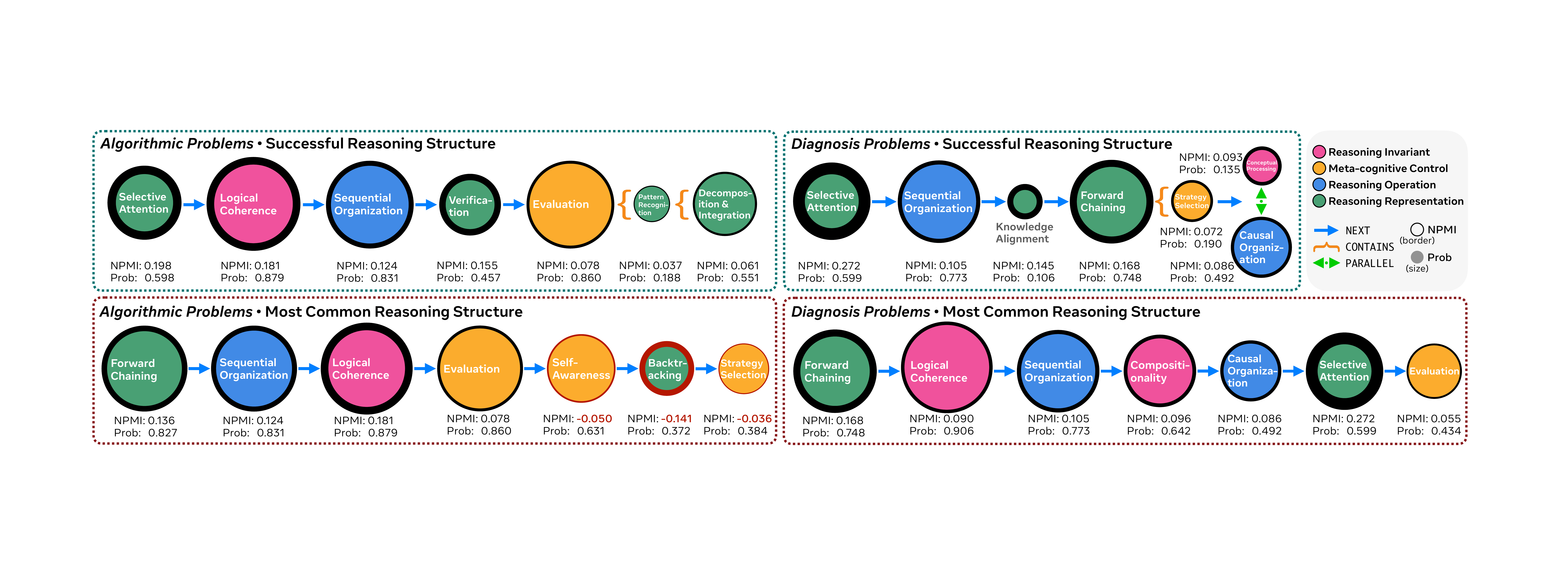}\vspace{-1mm}
    \caption{\small{Successful ($G^*$) and typical ($\hat{G}$) reasoning topologies on algorithmic and diagnosis-solution problems, with each element's association with success (NPMI) and presence rate (Prob.). Successful topologies begin with \Operation{selective attention}, and the typical topologies begin with \Operation{forward chaining}.}}
    \label{fig:transition_graph_example}\vspace{-2mm}
\end{figure}

\textbf{\new{Successful \newtwo{topologies} represent the problem before generating solutions.}} 
In both algorithmic (well-structured) and diagnosis-solution (\newtwo{intermediate}) problems, the successful \newtwo{topology} ($G^*$) begins with \Operation{selective attention} (Figure~\ref{fig:transition_graph_example}). In other words, successful traces identify relevant features and align the problem with domain constraints before constructing a solution. 
However, the typical reasoning \newtwo{topology} $\hat{G}$ has a strong tendency to start directly with \Operation{forward chaining}, \newthree{which is the first element in 36\% of algorithmic traces and 45\% of} \newthree{diagnosis-solution} \newthree{traces, more often than any other element}, then proceed to \Representation{sequential organization}, similar to their successful counterparts. This suggests that typical and successful model reasoning construct solutions in a similar way and differ mainly in whether they first represent the problem.
The typical \newtwo{topology} in algorithmic problems also ends with two elements associated with failure, \Operation{backtracking} (NPMI \new{$-0.173$}) and \Metacognitive{self-awareness} (\new{$-0.072$}). 
This pattern of starting with \Operation{selective attention} in successful traces echoes human behaviors, where expert problem solvers likewise represent a problem in terms of its underlying principles before solving it, whereas novices work from its surface features \citep{chi1981categorization}.

\textbf{Successful human and model \newtwo{topologies} are similar.}
On the matched corpus, the successful \newtwo{topologies} of humans and models share five of their seven elements (\Operation{forward chaining}, \Invariant{logical coherence}, \Operation{selective attention}, \Metacognitive{goal management}, and \Invariant{compositionality}). On each problem type, both begin with the same operation: \Operation{forward chaining} on story problems, and \Operation{selective attention} within their first two elements on \newtwo{intermediate and ill-structured} problems (App.~\ref{app:structures_matched}).
Because successful human and model reasoning converge on similar \newtwo{topologies} while typical model reasoning departs from them, these \newtwo{topologies} are a natural target for guiding models, which we test in Section~\ref{sec:steering}.

\begin{finding}
Models most often begin constructing a solution before representing the problem, whereas successful reasoning first identifies the problem's relevant features and constraints.
\end{finding}

\section{Eliciting Cognitive Reasoning Topologies via Test-time Reasoning Guidance}
\new{\label{sec:steering}}\vspace{-1mm}

\paragraph{Methodology.} 
Since successful human and model reasoning share much of the successful reasoning topologies for each problem type (Section~\ref{sec:behavior_structures}),
we now ask whether the extracted successful topologies can be used to guide models' reasoning process and improve task performance.
We propose \textbf{\textit{test-time reasoning guidance}}. Given a problem type, its successful topology $G^*$ is converted programmatically into a plain text string by listing its elements and each edge and edge type between them: ``\texttt{A $\rightarrow$ B}'' for \texttt{NEXT}, ``\texttt{A [contains] B}'' for \texttt{CONTAINS}, ``\texttt{A [parallel with] B}'' for \texttt{PARALLEL}.
The string and the problem type definition are then passed into a rewriter LLM (\texttt{Claude Sonnet 4.5}) and turned into natural language instructions explaining how to carry out each element on that problem type and how the elements relate. The reasoning guidance is then prepended to the problem text (Appendix~\ref{app:guidance}).

\vspace{-2mm}
\paragraph{Experimental setup.}
For each of the 15 open-weight text models and 11 problem types, we sample 50 problems, half of which the model originally answered correctly and half incorrectly, so that we can measure both whether guidance repairs failures and whether it harms problems the model already solves. 
Model responses with and without test-time reasoning guidance are graded by an LLM Judge (\texttt{GPT-4.1}, temp=0) against the same reference answer and judge prompt. 

\vspace{-2mm}
\paragraph{Results.}
Table~\ref{tab:model_performance} reports the change in accuracy under reasoning guidance. Models are ordered by their overall un-guided accuracy on the corpus (App.~\ref{app:accuracy_analysis}) and split into low-, mid-, and high-capability groups.

\begin{table*}[t]
\renewcommand{\arraystretch}{1.25}
\setlength{\tabcolsep}{4.5pt}
\centering
\caption{\small{Change in accuracy under test-time reasoning guidance (percentage points with bootstrapped standard errors). 
Guidance helps high-capability models most on ill-structured problems.}}\vspace{-2mm}
\label{tab:model_performance}
\resizebox{\textwidth}{!}{%
\footnotesize
\begin{tabular}{c|l|ccc|ccc|ccc|ccc|cc}
\toprule
 & & \multicolumn{3}{c|}{\textbf{Summary}} & \multicolumn{3}{c|}{\textbf{Well-structured}} & \multicolumn{3}{c|}{\textbf{Intermediate}} & \multicolumn{3}{c|}{\textbf{Ill-structured}} & \multicolumn{2}{c}{\textbf{Other}} \\
 & \textbf{Model} & \textbf{Avg.} & \textbf{Well} & \textbf{Ill} & \textbf{Algo} & \textbf{Story} & \textbf{Rule} & \textbf{Decision} & \textbf{Troub} & \textbf{Diag} & \textbf{Case} & \textbf{Design} & \textbf{Dilem} & \textbf{Log} & \textbf{Fact} \\
\hline
\multirow{6}{*}{\rotatebox{90}{\textbf{Low}}} & R1-Distill-Qwen-1.5B & \multicolumn{1}{>{\columncolor{figpink!8.8}[0pt][\tabcolsep]}c}{-9.5} & \cellcolor{figblue!2.5}+2.7 & \multicolumn{1}{>{\columncolor{figpink!9.9}[\tabcolsep][0pt]}c|}{-10.7} & \multicolumn{1}{>{\columncolor{figblue!3.7}[0pt][\tabcolsep]}c}{+4.0} & \cellcolor{figblue!11.1}+12.0 & \multicolumn{1}{>{\columncolor{figpink!7.4}[\tabcolsep][0pt]}c|}{-8.0} & \multicolumn{1}{>{\columncolor{figpink!29.6}[0pt][\tabcolsep]}c}{-32.0} & \cellcolor{figpink!14.8}-16.0 & \multicolumn{1}{>{\columncolor{figpink!3.7}[\tabcolsep][0pt]}c|}{-4.0} & \multicolumn{1}{>{\columncolor{figblue!1.9}[0pt][\tabcolsep]}c}{+2.0} & \cellcolor{figpink!13.0}-14.0 & \multicolumn{1}{>{\columncolor{figpink!18.5}[\tabcolsep][0pt]}c|}{-20.0} & \multicolumn{1}{>{\columncolor{figpink!9.3}[0pt][\tabcolsep]}c}{-10.0} & \cellcolor{figpink!16.7}-18.0 \\
 & DeepScaleR-1.5B & \multicolumn{1}{>{\columncolor{figpink!12.1}[0pt][\tabcolsep]}c}{-13.1} & \cellcolor{figpink!6.8}-7.3 & \multicolumn{1}{>{\columncolor{figpink!14.2}[\tabcolsep][0pt]}c|}{-15.3} & \multicolumn{1}{>{\columncolor{figpink!3.7}[0pt][\tabcolsep]}c}{-4.0} & \cellcolor{figpink!3.7}-4.0 & \multicolumn{1}{>{\columncolor{figpink!13.0}[\tabcolsep][0pt]}c|}{-14.0} & \multicolumn{1}{>{\columncolor{figpink!18.5}[0pt][\tabcolsep]}c}{-20.0} & \cellcolor{figpink!14.8}-16.0 & \multicolumn{1}{>{\columncolor{figpink!1.9}[\tabcolsep][0pt]}c|}{-2.0} & \multicolumn{1}{>{\columncolor{figpink!7.4}[0pt][\tabcolsep]}c}{-8.0} & \cellcolor{figpink!14.8}-16.0 & \multicolumn{1}{>{\columncolor{figpink!20.4}[\tabcolsep][0pt]}c|}{-22.0} & \multicolumn{1}{>{\columncolor{figpink!18.5}[0pt][\tabcolsep]}c}{-20.0} & \cellcolor{figpink!16.7}-18.0 \\
 & Hermes-3-Llama-3-8B & \multicolumn{1}{>{\columncolor{figblue!1.7}[0pt][\tabcolsep]}c}{+1.8} & \cellcolor{figpink!21.0}-22.7 & \multicolumn{1}{>{\columncolor{figblue!19.8}[\tabcolsep][0pt]}c|}{+21.3} & \multicolumn{1}{>{\columncolor{figpink!18.5}[0pt][\tabcolsep]}c}{-20.0} & \cellcolor{figpink!33.3}-36.0 & \multicolumn{1}{>{\columncolor{figpink!11.1}[\tabcolsep][0pt]}c|}{-12.0} & \multicolumn{1}{>{\columncolor{figblue!7.4}[0pt][\tabcolsep]}c}{+8.0} & \cellcolor{figblue!7.4}+8.0 & \multicolumn{1}{>{\columncolor{figblue!11.1}[\tabcolsep][0pt]}c|}{+12.0} & \multicolumn{1}{>{\columncolor{figblue!24.1}[0pt][\tabcolsep]}c}{+26.0} & \cellcolor{figblue!7.4}+8.0 & \multicolumn{1}{>{\columncolor{figblue!27.8}[\tabcolsep][0pt]}c|}{+30.0} & \multicolumn{1}{>{\columncolor{figpink!5.6}[0pt][\tabcolsep]}c}{-6.0} & \cellcolor{figblue!1.9}+2.0 \\
 & Olmo-3-7B-Think & \multicolumn{1}{>{\columncolor{figblue!6.2}[0pt][\tabcolsep]}c}{+6.7} & \cellcolor{figblue!16.0}+17.3 & \multicolumn{1}{>{\columncolor{figblue!9.9}[\tabcolsep][0pt]}c|}{+10.7} & \multicolumn{1}{>{\columncolor{figblue!14.8}[0pt][\tabcolsep]}c}{+16.0} & \cellcolor{figblue!13.0}+14.0 & \multicolumn{1}{>{\columncolor{figblue!20.4}[\tabcolsep][0pt]}c|}{+22.0} & \multicolumn{1}{>{\columncolor{figpink!5.6}[0pt][\tabcolsep]}c}{-6.0} & \cellcolor{figpink!5.6}-6.0 & \multicolumn{1}{>{\columncolor{figpink!7.4}[\tabcolsep][0pt]}c|}{-8.0} & \multicolumn{1}{>{\columncolor{figblue!9.3}[0pt][\tabcolsep]}c}{+10.0} & \cellcolor{figblue!0.0}0.0 & \multicolumn{1}{>{\columncolor{figblue!20.4}[\tabcolsep][0pt]}c|}{+22.0} & \multicolumn{1}{>{\columncolor{figblue!7.4}[0pt][\tabcolsep]}c}{+8.0} & \cellcolor{figblue!1.9}+2.0 \\
 & R1-Distill-Llama-8B & \multicolumn{1}{>{\columncolor{figpink!3.7}[0pt][\tabcolsep]}c}{-4.0} & \cellcolor{figpink!3.1}-3.3 & \multicolumn{1}{>{\columncolor{figpink!1.9}[\tabcolsep][0pt]}c|}{-2.0} & \multicolumn{1}{>{\columncolor{figpink!13.0}[0pt][\tabcolsep]}c}{-14.0} & \cellcolor{figblue!7.4}+8.0 & \multicolumn{1}{>{\columncolor{figpink!3.7}[\tabcolsep][0pt]}c|}{-4.0} & \multicolumn{1}{>{\columncolor{figpink!3.7}[0pt][\tabcolsep]}c}{-4.0} & \cellcolor{figpink!9.3}-10.0 & \multicolumn{1}{>{\columncolor{figpink!5.6}[\tabcolsep][0pt]}c|}{-6.0} & \multicolumn{1}{>{\columncolor{figpink!1.9}[0pt][\tabcolsep]}c}{-2.0} & \cellcolor{figpink!9.3}-10.0 & \multicolumn{1}{>{\columncolor{figblue!5.6}[\tabcolsep][0pt]}c|}{+6.0} & \multicolumn{1}{>{\columncolor{figpink!7.4}[0pt][\tabcolsep]}c}{-8.0} & \cellcolor{figblue!0.0}0.0 \\
 & \textit{Average} & \multicolumn{1}{>{\columncolor{figpink!3.3}[0pt][\tabcolsep]}c}{\textbf{-3.6}{\tiny$\,\pm$1.0}} & \cellcolor{figpink!2.5}-2.7{\tiny$\,\pm$2.0} & \multicolumn{1}{>{\columncolor{figblue!0.7}[\tabcolsep][0pt]}c|}{+0.8{\tiny$\,\pm$2.0}} & \multicolumn{1}{>{\columncolor{figpink!3.3}[0pt][\tabcolsep]}c}{-3.6{\tiny$\,\pm$3.3}} & \cellcolor{figpink!1.1}-1.2{\tiny$\,\pm$3.0} & \multicolumn{1}{>{\columncolor{figpink!3.0}[\tabcolsep][0pt]}c|}{-3.2{\tiny$\,\pm$3.8}} & \multicolumn{1}{>{\columncolor{figpink!10.0}[0pt][\tabcolsep]}c}{\textbf{-10.8}{\tiny$\,\pm$3.6}} & \cellcolor{figpink!7.4}\textbf{-8.0}{\tiny$\,\pm$3.6} & \multicolumn{1}{>{\columncolor{figpink!1.5}[\tabcolsep][0pt]}c|}{-1.6{\tiny$\,\pm$2.3}} & \multicolumn{1}{>{\columncolor{figblue!5.2}[0pt][\tabcolsep]}c}{+5.6{\tiny$\,\pm$3.4}} & \cellcolor{figpink!5.9}\textbf{-6.4}{\tiny$\,\pm$2.8} & \multicolumn{1}{>{\columncolor{figblue!3.0}[\tabcolsep][0pt]}c|}{+3.2{\tiny$\,\pm$4.1}} & \multicolumn{1}{>{\columncolor{figpink!6.7}[0pt][\tabcolsep]}c}{\textbf{-7.2}{\tiny$\,\pm$3.1}} & \cellcolor{figpink!5.9}-6.4{\tiny$\,\pm$3.7} \\
\hline
\multirow{6}{*}{\rotatebox{90}{\textbf{Mid}}} & R1-Distill-Qwen-7B & \multicolumn{1}{>{\columncolor{figpink!7.4}[0pt][\tabcolsep]}c}{-8.0} & \cellcolor{figblue!1.9}+2.0 & \multicolumn{1}{>{\columncolor{figpink!19.1}[\tabcolsep][0pt]}c|}{-20.7} & \multicolumn{1}{>{\columncolor{figblue!1.9}[0pt][\tabcolsep]}c}{+2.0} & \cellcolor{figblue!11.1}+12.0 & \multicolumn{1}{>{\columncolor{figpink!7.4}[\tabcolsep][0pt]}c|}{-8.0} & \multicolumn{1}{>{\columncolor{figpink!13.0}[0pt][\tabcolsep]}c}{-14.0} & \cellcolor{figpink!13.0}-14.0 & \multicolumn{1}{>{\columncolor{figpink!3.7}[\tabcolsep][0pt]}c|}{-4.0} & \multicolumn{1}{>{\columncolor{figpink!13.0}[0pt][\tabcolsep]}c}{-14.0} & \cellcolor{figpink!20.4}-22.0 & \multicolumn{1}{>{\columncolor{figpink!24.1}[\tabcolsep][0pt]}c|}{-26.0} & \multicolumn{1}{>{\columncolor{figpink!9.3}[0pt][\tabcolsep]}c}{-10.0} & \cellcolor{figblue!9.3}+10.0 \\
 & s1.1-32B & \multicolumn{1}{>{\columncolor{figblue!1.7}[0pt][\tabcolsep]}c}{+1.8} & \cellcolor{figblue!11.7}+12.7 & \multicolumn{1}{>{\columncolor{figblue!8.6}[\tabcolsep][0pt]}c|}{+9.3} & \multicolumn{1}{>{\columncolor{figblue!20.4}[0pt][\tabcolsep]}c}{+22.0} & \cellcolor{figblue!14.8}+16.0 & \multicolumn{1}{>{\columncolor{figblue!0.0}[\tabcolsep][0pt]}c|}{0.0} & \multicolumn{1}{>{\columncolor{figpink!9.3}[0pt][\tabcolsep]}c}{-10.0} & \cellcolor{figpink!13.0}-14.0 & \multicolumn{1}{>{\columncolor{figpink!1.9}[\tabcolsep][0pt]}c|}{-2.0} & \multicolumn{1}{>{\columncolor{figblue!11.1}[0pt][\tabcolsep]}c}{+12.0} & \cellcolor{figpink!1.9}-2.0 & \multicolumn{1}{>{\columncolor{figblue!16.7}[\tabcolsep][0pt]}c|}{+18.0} & \multicolumn{1}{>{\columncolor{figpink!16.7}[0pt][\tabcolsep]}c}{-18.0} & \cellcolor{figpink!1.9}-2.0 \\
 & Olmo-3-32B-Think & \multicolumn{1}{>{\columncolor{figblue!8.4}[0pt][\tabcolsep]}c}{+9.1} & \cellcolor{figblue!3.7}+4.0 & \multicolumn{1}{>{\columncolor{figblue!11.1}[\tabcolsep][0pt]}c|}{+12.0} & \multicolumn{1}{>{\columncolor{figpink!9.3}[0pt][\tabcolsep]}c}{-10.0} & \cellcolor{figblue!5.6}+6.0 & \multicolumn{1}{>{\columncolor{figblue!14.8}[\tabcolsep][0pt]}c|}{+16.0} & \multicolumn{1}{>{\columncolor{figblue!7.4}[0pt][\tabcolsep]}c}{+8.0} & \cellcolor{figblue!0.0}0.0 & \multicolumn{1}{>{\columncolor{figblue!14.8}[\tabcolsep][0pt]}c|}{+16.0} & \multicolumn{1}{>{\columncolor{figblue!3.7}[0pt][\tabcolsep]}c}{+4.0} & \cellcolor{figblue!0.0}0.0 & \multicolumn{1}{>{\columncolor{figblue!29.6}[\tabcolsep][0pt]}c|}{+32.0} & \multicolumn{1}{>{\columncolor{figblue!11.1}[0pt][\tabcolsep]}c}{+12.0} & \cellcolor{figblue!14.8}+16.0 \\
 & R1-Distill-Qwen-14B & \multicolumn{1}{>{\columncolor{figblue!0.7}[0pt][\tabcolsep]}c}{+0.7} & \cellcolor{figblue!3.7}+4.0 & \multicolumn{1}{>{\columncolor{figblue!2.5}[\tabcolsep][0pt]}c|}{+2.7} & \multicolumn{1}{>{\columncolor{figblue!3.7}[0pt][\tabcolsep]}c}{+4.0} & \cellcolor{figblue!5.6}+6.0 & \multicolumn{1}{>{\columncolor{figblue!1.9}[\tabcolsep][0pt]}c|}{+2.0} & \multicolumn{1}{>{\columncolor{figblue!1.9}[0pt][\tabcolsep]}c}{+2.0} & \cellcolor{figblue!0.0}0.0 & \multicolumn{1}{>{\columncolor{figpink!11.1}[\tabcolsep][0pt]}c|}{-12.0} & \multicolumn{1}{>{\columncolor{figblue!5.6}[0pt][\tabcolsep]}c}{+6.0} & \cellcolor{figblue!0.0}0.0 & \multicolumn{1}{>{\columncolor{figblue!1.9}[\tabcolsep][0pt]}c|}{+2.0} & \multicolumn{1}{>{\columncolor{figblue!3.7}[0pt][\tabcolsep]}c}{+4.0} & \cellcolor{figpink!5.6}-6.0 \\
 & R1-Distill-Qwen-32B & \multicolumn{1}{>{\columncolor{figblue!1.3}[0pt][\tabcolsep]}c}{+1.5} & \cellcolor{figblue!0.6}+0.7 & \multicolumn{1}{>{\columncolor{figblue!9.3}[\tabcolsep][0pt]}c|}{+10.0} & \multicolumn{1}{>{\columncolor{figpink!1.9}[0pt][\tabcolsep]}c}{-2.0} & \cellcolor{figblue!5.6}+6.0 & \multicolumn{1}{>{\columncolor{figpink!1.9}[\tabcolsep][0pt]}c|}{-2.0} & \multicolumn{1}{>{\columncolor{figpink!3.7}[0pt][\tabcolsep]}c}{-4.0} & \cellcolor{figblue!1.9}+2.0 & \multicolumn{1}{>{\columncolor{figpink!1.9}[\tabcolsep][0pt]}c|}{-2.0} & \multicolumn{1}{>{\columncolor{figblue!9.3}[0pt][\tabcolsep]}c}{+10.0} & \cellcolor{figblue!5.6}+6.0 & \multicolumn{1}{>{\columncolor{figblue!13.0}[\tabcolsep][0pt]}c|}{+14.0} & \multicolumn{1}{>{\columncolor{figpink!16.7}[0pt][\tabcolsep]}c}{-18.0} & \cellcolor{figblue!5.6}+6.0 \\
 & \textit{Average} & \multicolumn{1}{>{\columncolor{figblue!0.9}[0pt][\tabcolsep]}c}{+1.0{\tiny$\,\pm$1.0}} & \cellcolor{figblue!4.3}\textbf{+4.7}{\tiny$\,\pm$1.8} & \multicolumn{1}{>{\columncolor{figblue!2.5}[\tabcolsep][0pt]}c|}{+2.7{\tiny$\,\pm$2.1}} & \multicolumn{1}{>{\columncolor{figblue!3.0}[0pt][\tabcolsep]}c}{+3.2{\tiny$\,\pm$3.2}} & \cellcolor{figblue!8.5}\textbf{+9.2}{\tiny$\,\pm$2.7} & \multicolumn{1}{>{\columncolor{figblue!1.5}[\tabcolsep][0pt]}c|}{+1.6{\tiny$\,\pm$3.3}} & \multicolumn{1}{>{\columncolor{figpink!3.3}[0pt][\tabcolsep]}c}{-3.6{\tiny$\,\pm$3.5}} & \cellcolor{figpink!4.8}-5.2{\tiny$\,\pm$3.2} & \multicolumn{1}{>{\columncolor{figpink!0.7}[\tabcolsep][0pt]}c|}{-0.8{\tiny$\,\pm$3.2}} & \multicolumn{1}{>{\columncolor{figblue!3.3}[0pt][\tabcolsep]}c}{+3.6{\tiny$\,\pm$3.3}} & \cellcolor{figpink!3.3}-3.6{\tiny$\,\pm$3.5} & \multicolumn{1}{>{\columncolor{figblue!7.4}[\tabcolsep][0pt]}c|}{\textbf{+8.0}{\tiny$\,\pm$3.9}} & \multicolumn{1}{>{\columncolor{figpink!5.6}[0pt][\tabcolsep]}c}{-6.0{\tiny$\,\pm$3.7}} & \cellcolor{figblue!4.4}+4.8{\tiny$\,\pm$3.0} \\
\hline
\multirow{6}{*}{\rotatebox{90}{\textbf{High}}} & R1-Distill-Llama-70B & \multicolumn{1}{>{\columncolor{figblue!6.1}[0pt][\tabcolsep]}c}{+6.5} & \cellcolor{figblue!8.6}+9.3 & \multicolumn{1}{>{\columncolor{figblue!9.9}[\tabcolsep][0pt]}c|}{+10.7} & \multicolumn{1}{>{\columncolor{figblue!7.4}[0pt][\tabcolsep]}c}{+8.0} & \cellcolor{figblue!1.9}+2.0 & \multicolumn{1}{>{\columncolor{figblue!16.7}[\tabcolsep][0pt]}c|}{+18.0} & \multicolumn{1}{>{\columncolor{figblue!0.0}[0pt][\tabcolsep]}c}{0.0} & \cellcolor{figblue!1.9}+2.0 & \multicolumn{1}{>{\columncolor{figpink!3.7}[\tabcolsep][0pt]}c|}{-4.0} & \multicolumn{1}{>{\columncolor{figblue!11.1}[0pt][\tabcolsep]}c}{+12.0} & \cellcolor{figblue!0.0}0.0 & \multicolumn{1}{>{\columncolor{figblue!18.5}[\tabcolsep][0pt]}c|}{+20.0} & \multicolumn{1}{>{\columncolor{figpink!1.9}[0pt][\tabcolsep]}c}{-2.0} & \cellcolor{figblue!14.8}+16.0 \\
 & OpenThinker-32B & \multicolumn{1}{>{\columncolor{figblue!4.9}[0pt][\tabcolsep]}c}{+5.3} & \cellcolor{figblue!11.7}+12.7 & \multicolumn{1}{>{\columncolor{figblue!8.6}[\tabcolsep][0pt]}c|}{+9.3} & \multicolumn{1}{>{\columncolor{figpink!1.9}[0pt][\tabcolsep]}c}{-2.0} & \cellcolor{figblue!14.8}+16.0 & \multicolumn{1}{>{\columncolor{figblue!22.2}[\tabcolsep][0pt]}c|}{+24.0} & \multicolumn{1}{>{\columncolor{figpink!3.7}[0pt][\tabcolsep]}c}{-4.0} & \cellcolor{figpink!3.7}-4.0 & \multicolumn{1}{>{\columncolor{figpink!1.9}[\tabcolsep][0pt]}c|}{-2.0} & \multicolumn{1}{>{\columncolor{figblue!5.6}[0pt][\tabcolsep]}c}{+6.0} & \cellcolor{figblue!1.9}+2.0 & \multicolumn{1}{>{\columncolor{figblue!18.5}[\tabcolsep][0pt]}c|}{+20.0} & \multicolumn{1}{>{\columncolor{figpink!5.6}[0pt][\tabcolsep]}c}{-6.0} & \cellcolor{figblue!7.4}+8.0 \\
 & Qwen3-8B & \multicolumn{1}{>{\columncolor{figblue!1.7}[0pt][\tabcolsep]}c}{+1.8} & \cellcolor{figblue!1.9}+2.0 & \multicolumn{1}{>{\columncolor{figblue!6.8}[\tabcolsep][0pt]}c|}{+7.3} & \multicolumn{1}{>{\columncolor{figblue!0.0}[0pt][\tabcolsep]}c}{0.0} & \cellcolor{figblue!3.7}+4.0 & \multicolumn{1}{>{\columncolor{figblue!1.9}[\tabcolsep][0pt]}c|}{+2.0} & \multicolumn{1}{>{\columncolor{figblue!1.9}[0pt][\tabcolsep]}c}{+2.0} & \cellcolor{figblue!1.9}+2.0 & \multicolumn{1}{>{\columncolor{figblue!1.9}[\tabcolsep][0pt]}c|}{+2.0} & \multicolumn{1}{>{\columncolor{figblue!1.9}[0pt][\tabcolsep]}c}{+2.0} & \cellcolor{figblue!9.3}+10.0 & \multicolumn{1}{>{\columncolor{figblue!9.3}[\tabcolsep][0pt]}c|}{+10.0} & \multicolumn{1}{>{\columncolor{figpink!5.6}[0pt][\tabcolsep]}c}{-6.0} & \cellcolor{figpink!7.4}-8.0 \\
 & Qwen3-14B & \multicolumn{1}{>{\columncolor{figblue!7.4}[0pt][\tabcolsep]}c}{+8.0} & \cellcolor{figpink!0.6}-0.7 & \multicolumn{1}{>{\columncolor{figblue!16.0}[\tabcolsep][0pt]}c|}{+17.3} & \multicolumn{1}{>{\columncolor{figpink!3.7}[0pt][\tabcolsep]}c}{-4.0} & \cellcolor{figblue!0.0}0.0 & \multicolumn{1}{>{\columncolor{figblue!1.9}[\tabcolsep][0pt]}c|}{+2.0} & \multicolumn{1}{>{\columncolor{figblue!1.9}[0pt][\tabcolsep]}c}{+2.0} & \cellcolor{figblue!16.7}+18.0 & \multicolumn{1}{>{\columncolor{figblue!9.3}[\tabcolsep][0pt]}c|}{+10.0} & \multicolumn{1}{>{\columncolor{figblue!14.8}[0pt][\tabcolsep]}c}{+16.0} & \cellcolor{figblue!1.9}+2.0 & \multicolumn{1}{>{\columncolor{figblue!31.5}[\tabcolsep][0pt]}c|}{+34.0} & \multicolumn{1}{>{\columncolor{figblue!0.0}[0pt][\tabcolsep]}c}{0.0} & \cellcolor{figblue!7.4}+8.0 \\
 & Qwen3-32B & \multicolumn{1}{>{\columncolor{figblue!4.5}[0pt][\tabcolsep]}c}{+4.9} & \cellcolor{figblue!5.6}+6.0 & \multicolumn{1}{>{\columncolor{figblue!3.7}[\tabcolsep][0pt]}c|}{+4.0} & \multicolumn{1}{>{\columncolor{figblue!3.7}[0pt][\tabcolsep]}c}{+4.0} & \cellcolor{figblue!13.0}+14.0 & \multicolumn{1}{>{\columncolor{figblue!0.0}[\tabcolsep][0pt]}c|}{0.0} & \multicolumn{1}{>{\columncolor{figpink!3.7}[0pt][\tabcolsep]}c}{-4.0} & \cellcolor{figblue!7.4}+8.0 & \multicolumn{1}{>{\columncolor{figblue!14.8}[\tabcolsep][0pt]}c|}{+16.0} & \multicolumn{1}{>{\columncolor{figblue!5.6}[0pt][\tabcolsep]}c}{+6.0} & \cellcolor{figpink!3.7}-4.0 & \multicolumn{1}{>{\columncolor{figblue!9.3}[\tabcolsep][0pt]}c|}{+10.0} & \multicolumn{1}{>{\columncolor{figpink!5.6}[0pt][\tabcolsep]}c}{-6.0} & \cellcolor{figblue!9.3}+10.0 \\
 & \textit{Average} & \multicolumn{1}{>{\columncolor{figblue!4.9}[0pt][\tabcolsep]}c}{\textbf{+5.3}{\tiny$\,\pm$0.9}} & \cellcolor{figblue!5.4}\textbf{+5.9}{\tiny$\,\pm$1.6} & \multicolumn{1}{>{\columncolor{figblue!9.0}[\tabcolsep][0pt]}c|}{\textbf{+9.7}{\tiny$\,\pm$1.8}} & \multicolumn{1}{>{\columncolor{figblue!1.1}[0pt][\tabcolsep]}c}{+1.2{\tiny$\,\pm$2.6}} & \cellcolor{figblue!6.7}\textbf{+7.2}{\tiny$\,\pm$2.9} & \multicolumn{1}{>{\columncolor{figblue!8.5}[\tabcolsep][0pt]}c|}{\textbf{+9.2}{\tiny$\,\pm$2.8}} & \multicolumn{1}{>{\columncolor{figpink!0.7}[0pt][\tabcolsep]}c}{-0.8{\tiny$\,\pm$3.0}} & \cellcolor{figblue!4.8}+5.2{\tiny$\,\pm$3.5} & \multicolumn{1}{>{\columncolor{figblue!4.1}[\tabcolsep][0pt]}c|}{+4.4{\tiny$\,\pm$3.6}} & \multicolumn{1}{>{\columncolor{figblue!7.8}[0pt][\tabcolsep]}c}{\textbf{+8.4}{\tiny$\,\pm$2.8}} & \cellcolor{figblue!1.9}+2.0{\tiny$\,\pm$2.6} & \multicolumn{1}{>{\columncolor{figblue!17.4}[\tabcolsep][0pt]}c|}{\textbf{+18.8}{\tiny$\,\pm$3.7}} & \multicolumn{1}{>{\columncolor{figpink!3.7}[0pt][\tabcolsep]}c}{-4.0{\tiny$\,\pm$3.2}} & \cellcolor{figblue!6.3}\textbf{+6.8}{\tiny$\,\pm$2.8} \\
\hline
\multicolumn{2}{l|}{\textit{All models}} & \multicolumn{1}{>{\columncolor{figblue!0.8}[0pt][\tabcolsep]}c}{+0.9{\tiny$\,\pm$0.6}} & \cellcolor{figblue!2.4}\textbf{+2.6}{\tiny$\,\pm$1.1} & \multicolumn{1}{>{\columncolor{figblue!4.1}[\tabcolsep][0pt]}c|}{\textbf{+4.4}{\tiny$\,\pm$1.2}} & \multicolumn{1}{>{\columncolor{figblue!0.2}[0pt][\tabcolsep]}c}{+0.3{\tiny$\,\pm$1.8}} & \cellcolor{figblue!4.7}\textbf{+5.1}{\tiny$\,\pm$1.8} & \multicolumn{1}{>{\columncolor{figblue!2.3}[\tabcolsep][0pt]}c|}{+2.5{\tiny$\,\pm$1.9}} & \multicolumn{1}{>{\columncolor{figpink!4.7}[0pt][\tabcolsep]}c}{\textbf{-5.1}{\tiny$\,\pm$2.2}} & \cellcolor{figpink!2.5}-2.7{\tiny$\,\pm$2.3} & \multicolumn{1}{>{\columncolor{figblue!0.6}[\tabcolsep][0pt]}c|}{+0.7{\tiny$\,\pm$2.0}} & \multicolumn{1}{>{\columncolor{figblue!5.4}[0pt][\tabcolsep]}c}{\textbf{+5.9}{\tiny$\,\pm$1.7}} & \cellcolor{figpink!2.5}-2.7{\tiny$\,\pm$1.9} & \multicolumn{1}{>{\columncolor{figblue!9.3}[\tabcolsep][0pt]}c|}{\textbf{+10.0}{\tiny$\,\pm$2.5}} & \multicolumn{1}{>{\columncolor{figpink!5.3}[0pt][\tabcolsep]}c}{\textbf{-5.7}{\tiny$\,\pm$2.1}} & \cellcolor{figblue!1.6}+1.7{\tiny$\,\pm$1.8} \\
\bottomrule
\end{tabular}
}
\end{table*}

\vspace{-1mm}
\textbf{Guidance is most helpful on ill-structured problems.} 
Across all models, reasoning guidance is more helpful on ill-structured problems ($+4.4_{\pm1.2}$) than well-structured problems ($+1.6_{\pm1.1}$).
Especially on high-capability models, the gain is largest on the three ill-structured types (\newtwo{$+9.7_{\pm 1.8}$}), led by dilemmas ($+18.8_{\pm 3.7}$) and case analysis ($+8.4_{\pm 2.8}$), and smaller on well-structured problems ($+5.9_{\pm 1.6}$). 
Cognitive elements are more predictive of success on less structured problems (Figure~\ref{fig:presence_ppmi_heatmaps_v5}, right), and on ill-structured problems human \newthree{meta-reasoning} increases with uncertainty while model \newthree{meta-reasoning} follows verifiability (Figure~\ref{fig:prolific_adaptation_regulation}b--e), so guidance helps most where Section~\ref{sec:results} locates the gap. 

\vspace{-1mm}
\new{\textbf{The benefit depends on model capability.} The effect of guidance grows with a model's overall accuracy ($p<10^{-15}$). For high-capability models, guidance raises accuracy from 59.0\% to 64.3\% ($+5.3_{\pm 0.9}$). Mid-capability models are essentially unchanged ($+1.0_{\pm 1.0}$), and low-capability models lose accuracy ($-3.6_{\pm 1.0}$), most of all \texttt{DeepScaleR-1.5B} ($-13.1$) and \texttt{R1-Distill-Qwen-1.5B} ($-9.5$). 
Following a multi-phase reasoning plan while solving the problem is itself demanding, and models without the capacity to do both lose more from the disruption than they gain from the topology.}

\vspace{-1mm}
\begin{finding}
Capable models possess the elements that predict success but do not deploy them naturally, and that a topology derived automatically from successful traces can elicit them.
\end{finding}

\section{Opportunities and Challenges}

Models exhibit nearly every element of the taxonomy but differ from humans in how they use them, and guidance derived from successful traces improves capable models on ill-structured problems in-context.
Describing reasoning by its cognitive elements thus reveals differences that accuracy alone does not, and points to where model reasoning could improve.
We use the taxonomy to compare these findings with the elements targeted by LLM research (Section~\ref{sec:research_design}) and to identify open problems on which cognitive science can make testable predictions and contributions on model reasoning (Section~\ref{sec:open_challenges}).

\subsection{\new{Cognitive Element Considerations in LLM Research Design}}
\label{sec:research_design}

\par To contextualize the behavioral presence observed in Section \ref{sec:behavior_distribution}, we examine how contemporary LLM research conceptualizes ``reasoning'' as a design choice. With growing interest in the development and analysis of models with ``diverse and strong reasoning'' abilities, we aim to understand what cognitive elements are currently supported and which remain underexplored. For this analysis, we scrape arXiv papers using keyword-based queries applied to titles and abstracts. We search using general reasoning-related keywords (e.g., ``LLM reasoning,'' ``LLM cognitive behaviors,'' ``LLM thinking'') and for each capability in our taxonomy we add an additional query of the form ``LLM \textit{<capability>},'' retrieving the top ten papers for each general query (in order to maximize precision) and 100 papers for each behavior-specific query. Because each element query contributes a fixed number of papers, the percentages below describe this retrieved pool and are not a representative sample of the field.

We annotate a subset of papers using two human annotators and \texttt{GPT-4.1} (all three achieve moderate agreement $ICC_{3k}=0.593$; we iteratively improve annotation prompts based on human feedback), and then the entire set of \newthree{1,598} papers using \texttt{GPT-4.1} with the iterated prompts. %
\new{For each paper--element pair, we record whether the element is \textit{present} (a deliberate and significant design decision, e.g., in evaluation objectives or architecture), \textit{partially present} (considered in the design, e.g., through dataset structure or demonstration style, but not the primary focus), or absent.}

\begin{wrapfigure}{R}{0.5\textwidth}%
    \centering\vspace{-4mm}
    \new{\includegraphics[width=\linewidth]{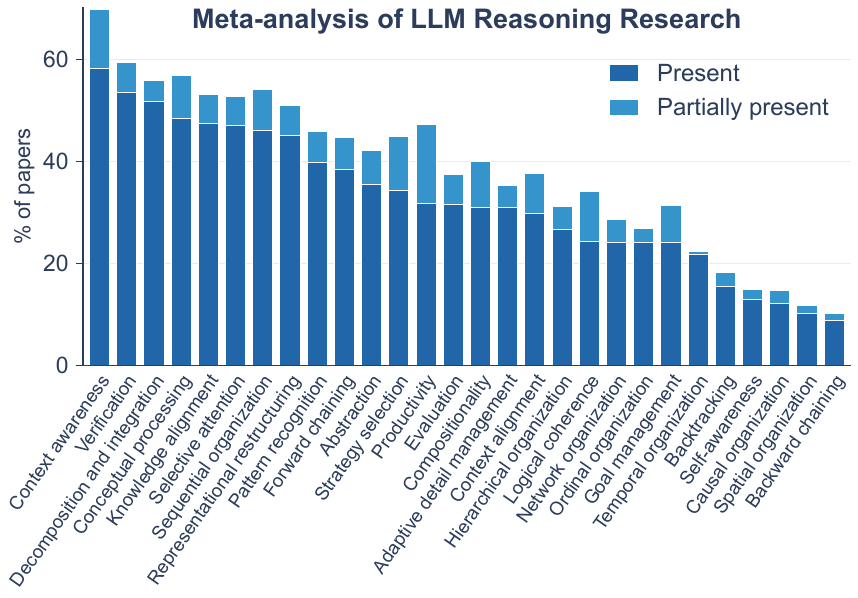}}\vspace{-2.6mm}
    \caption{\small{\% of LLM reasoning papers in which each cognitive element is present or partially present in the research design. Research concentrates on context awareness, verification, and decomposition, but rarely targets self-awareness, causal/spatial reasoning, or backward chaining.}}\vspace{-6mm}
    \label{fig:paper_behaviors}
\end{wrapfigure}%

Results from this meta-analysis reveal substantial imbalance in the types of cognitive elements emphasized across LLM reasoning research (\autoref{fig:paper_behaviors}). The elements most often present in the research design are context awareness (58\% of papers), verification (54\%), decomposition and integration (52\%), conceptual processing (49\%), knowledge alignment (48\%), and sequential organization (46\%). In contrast, elements enabling flexible problem-solving receive far less attention, with temporal organization present in 22\% of papers, self-awareness in 13\%, causal organization in 12\%, spatial organization in 10\%, and backward chaining in 9\%.

Current LLM reasoning research operates within a narrow conceptual vocabulary, \newthree{concentrating on elements that can be evaluated on structured, verifiable problems, such as verification, decomposition, and sequential organization, and rarely targeting self-awareness, causal, temporal, and spatial organization.} 
The research focus trends are partially reflected in the LLM reasoning behavior observed in current models' reasoning traces. For example, \Operation{verification} is a heavily optimized \newthree{cognitive element}, which could contribute to its abundance in well-structured problems and its drop in ill-structured problems. On the other hand, meta-cognitive \newthree{controls} such as \Metacognitive{goal management} (24\%) \newthree{and}
\Metacognitive{self-awareness} (13\%) are understudied, \newthree{and both differ between human and model reasoning on ill-structured problems (Section~\ref{sec:problem_type}).}
Expanding theoretical foundations beyond step-by-step decomposition toward richer cognitive taxonomies may be essential for developing more human-like and generalizable reasoning capabilities.

\subsection{\new{Open Challenges}}\label{sec:open_challenges}

In this paper, we introduce a cognitively grounded framework for analyzing reasoning traces via their use of cognitive elements. Differences in model versus human, correct versus incorrect, and well-structured versus ill-structured reasoning pairs all open up avenues for further research. 
Our proposed taxonomy introduces a shared vocabulary, through which insights from cognitive science research on human reasoning and AI research on model reasoning can more readily inform each other.

\paragraph{Formulating ill-structured problems.}
One needs to reason about how to represent and formulate the problem before solving it \citep{simon1973structure}. 
The formulation process is almost trivial in well-structured problems; yet for ill-structured problems, the reasoner first decides which goals, constraints, and unknowns matter, and much of the difficulty lies in this step \citep{kaplan1990search}.
Most research on LLM reasoning uses problems that are already formulated, and work on formulation has focused on translating well-specified problems into formal programs \citep{zhou2025autoformulating}.
Our results suggest that models often skip this formulation step.
On less structured problems, successful reasoning starts by identifying the relevant features of the problem with \Operation{selective attention}, while models most often start generating a solution with \Operation{forward chaining} right away (Section~\ref{sec:behavior_structures}).
In both humans and models, problem formulation has been hard to study because it does not produce an answer, and formulation correctness cannot be easily disentangled from downstream reasoning.
Our analysis framework makes the problem formulation step visible and hence measurable in any given reasoning trace.
Problem formulation capability would empower LLMs to solve a much broader range of problems.

\paragraph{Learning to select strategies by recognizing problem type.}
Human experts internalize and speed up the strategy selection process by recognizing the deep structure of a problem through fast pattern recognition, and the structure informs the appropriate strategy \citep{chase1973perception,klein1998sources}.
Models, however, are able to follow a given successful reasoning topology but rarely choose the correct one independently (Section~\ref{sec:steering}).
LLM research has approached strategy selection by letting strategies emerge through reinforcement learning \citep{el2025competitive}, by prompting models to choose among reasoning methods \citep{zhou2024selfdiscover,gao2024meta}, or by training the choice directly \citep{liu2024smart}, but these approaches leave open whether the choice rests on a problem's structure or on its surface.
Our problem-type based analysis offers a way to evaluate strategy selection in both humans and models.
Further, probing model's internal representations of problem type, and aligning it to the successful strategy can instill expert-like strategy selection and turn the gains of test-time reasoning guidance into a capability of the model.

\paragraph{\newthree{Preserving strategy diversity during training.}}
Children use several strategies for the same kind of problem, and those who use more of them are more likely to discover better ones \citep{siegler1996emerging,Shrager1998-oq}.
Yet reinforcement learning improves accuracy at the cost of narrowing the range of solutions a model produces \citep{yue2025does}.
Work on exploration measures this narrowing over tokens and rely heavily on the specific choice of embeddings \citep{li2025jointlyreinforcingdiversityquality}, but the developmental evidence suggests that what matters for discovery is diversity over strategies \citep{lee2026measuringstrategyphrasinggap}.
Our results are consistent with such narrowing, as models most often follow a topology that starts solving before representing the problem (Section~\ref{sec:behavior_structures}).
Our taxonomy can measure strategy diversity throughout training and correlate it to exploration. 
Using the proposed metrics such as element presence rate and association with success (NPMI) in reward shaping, or seeding trajectories with rarely used elements, could preserve reasoning diversity and improve generalization.

\paragraph{\newthree{A developmental account of reasoning in models.}}
In what order do the elements of reasoning emerge, and what drives each step?
Developmental psychology studies this in children from snapshots at different ages.
Analyzing model checkpoints at different training stages and conditions using our taxonomy, \newthree{as microgenetic studies do for children \citep{siegler1996emerging},} can tell us whether each element comes from pretraining, distillation, or reinforcement learning, or when model meta-reasoning starts to differ from human meta-reasoning.
For computer scientists, knowing which stage of training produces each element gives insight into targeted capability improvement; 
for cognitive scientists, comparing developmental model reasoning with developmental human reasoning would inform whether the order in which reasoning develops in children reflects the structure of reasoning itself or the course of human development.

\paragraph{Reasoning faithfulness and monitorability through cognitive operations.}
Reasoning traces are used to evaluate and monitor models, but it is unclear whether a trace reflects the computation that produced the answer \citep{turpin2023language,chen2025reasoning}.
Existing tests answer this for a whole trace at once, for example by truncating the trace and checking whether the answer changes \citep{lanham2023measuring}, so they cannot tell which parts of the trace the model relied on.
Our results show that elements already differ in their value for monitoring, as \Operation{verification} tracks correctness while \Invariant{logical coherence} appears in almost every trace regardless of outcome (Section~\ref{sec:success}).
Cognitive science faced this question with human think-aloud data, where it is likewise unclear whether what people say reflects how they actually reasoned, and addressed it by coding each verbalization as an operation and checking it against an independent record of the process \citep{newell1972human,russo1989validity}.
Our taxonomy allows measuring of faithfulness of each cognitive element, such as changing the outcome of a \Operation{verification} step, and observing whether the answer changes accordingly.
Knowing which elements are faithful would link corresponding spans to the model's actual computation, making model reasoning truly monitorable.

\section*{Acknowledgments}
We are grateful for the helpful discussions with Kerem Oktar about the cognitive taxonomy design. 
We wish to express our sincere gratitude to 
Anshul Nasery, 
Kshitish Ghate, 
Divyansh Pareek, 
Moe Kayali, 
Runjia Li, 
Harshita Chopra, 
Mihir Kavishwar, 
Ishika Agarwal, 
and Amruta Parulekar
for their help in collecting human reasoning traces.
This material is based upon work supported by the National Science Foundation under Grant No.~IIS2554007.
This research was developed in part with funding from the Defense Advanced Research Projects Agency's (DARPA) SciFy program (Agreement No. HR00112520300). The views expressed are those of the author and do not reflect the official policy or position of the Department of Defense or the U.S.~Government. 
This material is based in part upon work supported by the Defense Advanced Research Projects Agency and the Air Force Research Laboratory, contract number(s): FA8650-23-C-7316. Any opinions, findings and conclusions, or recommendations expressed in this material are those of the author(s) and do not necessarily reflect the views of AFRL or DARPA.
This research was supported by Coefficient Giving, the University of Washington Population Health Initiative, Amazon Health, the UW+Amazon Science Hub, and the Meta AIM program.

\section*{Author Contributions}
\begin{itemize}[itemsep=0pt,topsep=-2pt,leftmargin=16pt]
    \item Cognitive Element Taxonomy: SSL, PK, HW, JL, DL
    \item Dataset Construction. Text + Audio + Vision: SSL, PK, OA; Human: PK, SSL, JL, DL

    \item Cognitive Element Annotation \& Evaluation: SSL, PK, JL, DL
    \item Problem Structure Type Classification: SSL, PK
    \item Cognitive Element Analysis. LLMs: PK, SSL, JL; Human vs. LLMs: SSL, PK, JL, DL, HW, SC

    \item Reasoning \newtwo{Topologies}: PK, SSL, JL
    \item Eliciting Reasoning \newtwo{Topologies}: PK, SSL
    \item LLM Papers Meta-analysis: PK, SSL, JL
    \item Open Challenges: SSL

    \item Advising: TG, MK, JH, AC, YT
\end{itemize}

\bibliography{tmlr}
\bibliographystyle{tmlr}

\appendix
\section{Appendix}
\renewcommand{\addcontentsline}[3]{}%

\subsection{Typology of Problems}\label{app:problem_types}

We classify cognitive tasks using an extended version of the problem-solving taxonomy of \newthree{\citet{jonassen2000toward}}. Jonassen's framework characterizes problems along a continuum from well-structured (clear goals, known solution paths, predictable outcomes) to ill-structured (ambiguous goals, multiple solution paths, uncertain outcomes), organizing problems into 11 types based on their structural properties and cognitive demands.

\subsubsection{Problem Type Definitions}

Following \citet{jonassen2000toward}, we define problem-solving as a goal-directed cognitive activity that transforms an initial state into a desired goal state through systematic reasoning. \new{Ten of the 11 problem-solving types lie on the structuredness continuum, which we divide into well-structured, intermediate, and ill-structured problems as in Table~\ref{tab:model_performance}. Logical problems fall outside the continuum.}

\textbf{Well-Structured Problems}
\begin{enumerate}[itemsep=0pt,topsep=-2pt,leftmargin=16pt]
    \item \textbf{Algorithmic}: Fixed procedures applied to similar variable sets, producing correct answers through prescribed methods (e.g., solve quadratic equations, convert temperature units)
    \item \textbf{Story Problems}: Mathematical or scientific problems embedded in narrative contexts, requiring extraction of values and formula application (e.g., distance-rate-time problems)
    \item \textbf{Rule-Using}: Procedural processes constrained by rules that allow multiple valid approaches to system-constrained answers (e.g., database searching, theorem proving, recipe modification)
\end{enumerate}

\textbf{\new{Intermediate Problems}}
\begin{enumerate}[resume,itemsep=0pt,topsep=-2pt,leftmargin=16pt]
    \item \textbf{Decision-Making}: Selecting and justifying one option from a finite set of alternatives, weighing benefits and limitations (e.g., college selection, route planning, benefits package choice)
    \item \textbf{Troubleshooting}: Diagnosing faults in malfunctioning systems by generating and testing hypotheses (e.g., car won't start, network is down, code debugging)
    \item \textbf{Diagnosis-Solution}: Extending beyond fault identification to recommend and evaluate treatment options (e.g., medical diagnosis and treatment, identifying and treating lawn problems)
\end{enumerate}

\textbf{Ill-Structured Problems}
\begin{enumerate}[resume,itemsep=0pt,topsep=-2pt,leftmargin=16pt]
    \item \textbf{Strategic Performance}: Real-time execution of complex tactics while maintaining situational awareness under competing demands (e.g., flying aircraft, teaching live classes, managing portfolios during trading) \newthree{(no problem in our corpus is of this type)}
    \item \textbf{Case Analysis}: Analyzing complex scenarios with multiple stakeholders and perspectives, arguing positions in detail-rich situations with ill-defined goals (e.g., business cases, legal judgments, policy recommendations)
    \item \textbf{Design}: Creating new artifacts or systems that satisfy functional requirements, with solutions evaluated as better or worse rather than correct or incorrect (e.g., bridge design, curriculum development, marketing campaigns)
    \item \textbf{Dilemma}: Reconciling contradictory positions with no satisfactory solution that serves all perspectives (e.g., abortion policy, international conflicts, wealth redistribution)
\end{enumerate}

\textbf{\new{Outside the Continuum}}
\begin{enumerate}[resume,itemsep=0pt,topsep=-2pt,leftmargin=16pt]
    \item \textbf{Logical}: Abstract reasoning puzzles with optimal solutions and minimal context (e.g., Tower of Hanoi, river crossing puzzles)
\end{enumerate}

\subsubsection{Extension of Jonassen's Taxonomy}

We extend the original 11-category framework with two additional categories to capture tasks outside the goal-directed transformation paradigm:

\begin{enumerate}[resume,itemsep=0pt,topsep=-2pt,leftmargin=16pt]
    \item \textbf{Factual Recall}: Retrieving stored knowledge without requiring reasoning or problem-solving (e.g., ``What is photosynthesis?'' or ``List the causes of WWI'')
    \item \textbf{Creative/Expressive}: Generating novel content judged by originality or aesthetic quality rather than convergence to a predetermined solution (e.g., ``Write a poem'' or ``Draw how you feel'')
\end{enumerate}

\newthree{This yields a 13-category typology, of which 12 types occur in our datasets.}

\textbf{Classification Methodology}
\new{Each problem is classified by \texttt{o4-mini} (high reasoning effort), using detailed annotation guidelines based on the structural criteria of \citet{jonassen2000toward}, which are goal clarity, solution determinacy, and domain constraints.}

\textbf{Key Distinctions}
Several problem types share surface similarities but differ fundamentally in their cognitive demands:

\begin{itemize}[itemsep=0pt,topsep=-2pt,leftmargin=16pt]
    \item \textbf{Troubleshooting vs. Diagnosis-Solution}: Troubleshooting identifies faults; diagnosis-solution both identifies and treats

    \item \textbf{Story Problem vs. Factual Recall}: Story problems require calculation despite narrative framing; factual recall simply explains information

    \item \textbf{Decision-Making vs. Dilemma}: Decision-making has acceptable solutions; dilemmas have no satisfactory resolution for all parties

    \item \textbf{Design vs. Creative/Expressive}: Design has functional requirements and constraints; creative/expressive tasks involve pure expression

    \item \textbf{Algorithmic vs. Design}: Algorithms have one correct procedure; design problems have multiple valid approaches with better/worse solutions
\end{itemize}

\subsection{Length Normalization of Presence Scores}\label{app:length}
On the matched corpus, model traces are longer than human traces (median 475 vs.\ 326 words), and longer traces exhibit more elements (Spearman $\rho=.64$ for humans and $.76$ for models between total presence score and length). We therefore normalize each presence score by the logarithm of the trace's length:
\begin{equation*}
s'_{t,e} = s_{t,e}\,\frac{\log w^*}{\log w_t},
\end{equation*}
where $s_{t,e}\in\{0,1,2\}$ is the presence score of element $e$ in trace $t$, $w_t$ is the number of words in $t$, and $w^*=368$ is the median length over all traces, so a trace of median length keeps its score.
\newthree{Figure~\ref{fig:prolific_heatmap} reports the mean of $s'_{t,e}$ over the correct, incorrect, and all traces of each reasoner, and Figure~\ref{fig:prolific_adaptation_regulation}(b--e) reports it by problem structure for both corpora, with the same $w^*$.}
\newthree{Figure~\ref{fig:prolific_adaptation_regulation}(a) and the counts of elements present in Section~\ref{sec:problem_type} multiply each trace's count by the same factor $\log w^*/\log w_t$.}
\newthree{The length-normalized presence rates in Figures~\ref{fig:model_behavior_presence} (left) and~\ref{fig:presence_ppmi_heatmaps_v5} (left) weight each trace in which the element is clearly present by this factor, so a rate can exceed 1 when most traces are shorter than $w^*$.}
\newthree{The comparisons of presence between humans and models and between correct and incorrect traces on the matched corpus use $s'_{t,e}$.}
We normalize by log length rather than length because presence grows with diminishing returns: a presence score cannot exceed 2, and across length quintiles of human traces (median 96 to 708 words) each doubling of length adds a similar amount to the total presence score. Log length also explains more of the variation in total presence than raw length ($R^2=.41$ vs.\ $.35$ for humans, $.58$ vs.\ $.06$ for models).

\subsection{\new{Pilot Human Traces and Prompt Development}}\label{app:human_pilot}

We collected a pilot set of human reasoning traces \new{to develop the annotation prompts.} We recruited 18 human participants to solve a small subset of the \texttt{GeneralThought} dataset while recording their reasoning. Reasoning was recorded using a think-aloud protocol in which they recorded their verbal reasoning (later transcribed \newthree{automatically} \citep{evernote_ai_nodate}). Since some tasks require domain-specific facts or state tracking, we allow participants to use \textit{tools}, including web search and note-taking. In such cases, participants were instructed to verbalize the tool usage, e.g., speaking the search keyword aloud.
We collected a pilot set of human reasoning traces \new{to develop the annotation prompts.} We recruited 18 human participants to solve a small subset of the \texttt{GeneralThought} dataset while recording their reasoning. Reasoning was recorded using a think-aloud protocol in which they recorded their verbal reasoning (later transcribed with \cite{evernote_ai_nodate}). Since some tasks require domain-specific facts or state tracking, we \newthree{allowed} participants to use \textit{tools}, including web search and note-taking. In such cases, participants were instructed to verbalize the tool usage, e.g., speaking the search keyword aloud.

\new{Two human annotators scored each pilot trace on the three-level rubric (0 = absent, 1 = partially present, 2 = present) for each of the 28 elements, and we kept the lower of the two scores, so that an element counts as present only when both annotators judge it present. These annotations served as the reference for developing the span annotation prompts over several rounds, with the same annotators reviewing the model's annotations and rationales in each round.}

\subsection{\new{Human Think-aloud Reasoning Collection}}\label{app:prolific}
\new{\textbf{Problems.} We wrote 30 problems for the study: ten story problems, ten diagnosis-solution problems, and ten case analysis problems, with five problems at each of two difficulty levels within each type. Solving the problems requires no domain expertise, and each problem's type is set by design. \newtwo{Table~\ref{tab:prolific_traces} shows one of the diagnosis-solution problems.}}

\new{\textbf{Participants and procedure.} We recruited 240 participants on Prolific under an IRB-approved protocol. Each participant solved between 1 and 13 problems (5.7 on average), assigned at random and balanced across difficulty and problem type, and thought aloud while solving each one \citep{ericsson1980verbal}. The recordings were transcribed automatically \citep{evernote_ai_nodate}. Each problem received between 27 and 62 transcripts (median 45).}

\new{\textbf{Exclusions.} Of the 1,365 transcripts, we excluded 19 that contain no reasoning attempt: 13 with three words or fewer and 6 not in English. The remaining 1,346 traces come from 236 participants, 215 of whom solved problems of all three types, and have a median length of \newthree{326} words (interquartile range 167 to \newthree{525}).}

\new{\textbf{Correctness.} We judge each trace's answer against a reference answer with the pairwise procedure of Section~\ref{sec:problem_eval}, using \texttt{GPT-4.1} as the judge with five votes. A trace is correct if the majority vote rates it better than or equivalent to the reference and incorrect if it rates it worse. Of the 1,346 human traces, 507 are correct, 836 are incorrect, and 3 receive an ambiguous decision and are left out of the correct-incorrect comparisons. Model traces on the same problems (one from each of 15 text models) are judged the same way.}

\subsection{\newtwo{Element Differences between Correct and Incorrect Traces}}\label{app:correctness_by_element}
\newtwo{Table~\ref{tab:correctness_by_element} reports, for each element on the matched corpus, the difference between correct and incorrect traces in two measures: the mean length-normalized presence score $s'_{t,e}$ (Appendix~\ref{app:length}), whose means for each outcome appear in Figure~\ref{fig:prolific_heatmap}, and the mean allocation share, the element's presence score divided by the sum of presence scores over all 28 elements. Each difference is tested with a Mann--Whitney $U$ test, Benjamini--Hochberg corrected across the 28 elements within each group and measure.
In presence, the differences of the two groups correlate across elements ($r=0.62$). Twelve elements are significantly more present in correct traces of both groups, led by \Operation{forward chaining}, \Invariant{compositionality}, \Operation{selective attention}, and \Operation{decomposition and integration}, and \Representation{network organization} is significantly less present in correct traces of both.
In allocation, the two groups agree less ($r=0.29$). Incorrect human traces give a larger share to \Metacognitive{self-awareness} (4.5 percentage points) and \Metacognitive{context awareness} (2.5), while incorrect model traces give a smaller share to \Operation{verification} (2.1) and \Metacognitive{evaluation} (1.7). \Invariant{Logical coherence} takes a larger share of correct human traces and a smaller share of correct model traces.}

\begin{table}[h]
\centering
\caption{\newtwo{Difference between correct and incorrect traces (correct minus incorrect) in each element's length-normalized presence score and allocation share (percentage points) on the matched corpus (bold: BH-corrected $q<0.05$); in both groups, correct traces contain more forward chaining, compositionality, selective attention, and decomposition.}}
\label{tab:correctness_by_element}
\resizebox{0.64\textwidth}{!}{%
\begin{tabular}{lrrrr}
\toprule
 & \multicolumn{2}{c}{Presence} & \multicolumn{2}{c}{Allocation (pp)} \\
\cmidrule(lr){2-3}\cmidrule(lr){4-5}
Element & Human & Model & Human & Model \\
\midrule
\multicolumn{5}{l}{\textit{Reasoning invariants}} \\
\quad Logical coherence & $\mathbf{+0.79}$ & $+0.17$ & $\mathbf{+2.40}$ & $\mathbf{-1.06}$ \\
\quad Compositionality & $\mathbf{+0.63}$ & $\mathbf{+0.50}$ & $\mathbf{+2.63}$ & $\mathbf{+0.63}$ \\
\quad Productivity & $-0.05$ & $+0.11$ & $\mathbf{-1.49}$ & $\mathbf{-0.31}$ \\
\quad Conceptual processing & $\mathbf{+0.14}$ & $\mathbf{+0.14}$ & $-0.20$ & $\mathbf{-0.53}$ \\
\midrule
\multicolumn{5}{l}{\textit{Meta-cognitive controls}} \\
\quad Self-awareness & $-0.06$ & $\mathbf{+0.33}$ & $\mathbf{-4.51}$ & $+0.48$ \\
\quad Context awareness & $\mathbf{-0.11}$ & $-0.08$ & $\mathbf{-2.46}$ & $\mathbf{-0.70}$ \\
\quad Strategy selection & $+0.04$ & $+0.11$ & $\mathbf{-0.93}$ & $\mathbf{-0.34}$ \\
\quad Goal management & $\mathbf{+0.28}$ & $\mathbf{+0.26}$ & $\mathbf{+0.31}$ & $-0.33$ \\
\quad Evaluation & $\mathbf{+0.13}$ & $\mathbf{+0.60}$ & $-0.43$ & $\mathbf{+1.71}$ \\
\midrule
\multicolumn{5}{l}{\textit{Reasoning representations}} \\
\quad Sequential organization & $\mathbf{+0.65}$ & $\mathbf{+0.32}$ & $\mathbf{+2.29}$ & $-0.37$ \\
\quad Hierarchical organization & $\mathbf{+0.20}$ & $\mathbf{+0.19}$ & $\mathbf{+0.56}$ & $-0.01$ \\
\quad Network organization & $\mathbf{-0.13}$ & $\mathbf{-0.18}$ & $\mathbf{-1.33}$ & $\mathbf{-1.58}$ \\
\quad Ordinal organization & $\mathbf{+0.21}$ & $+0.16$ & $-0.17$ & $\mathbf{0.00}$ \\
\quad Causal organization & $\mathbf{+0.11}$ & $+0.05$ & $\mathbf{-1.40}$ & $\mathbf{-0.77}$ \\
\quad Temporal organization & $\mathbf{+0.25}$ & $\mathbf{+0.19}$ & $-0.21$ & $\mathbf{-0.06}$ \\
\quad Spatial organization & $-0.02$ & $+0.04$ & $-0.18$ & $\mathbf{+0.14}$ \\
\midrule
\multicolumn{5}{l}{\textit{Reasoning operations}} \\
\quad Context alignment & $-0.04$ & $\mathbf{-0.10}$ & $\mathbf{-0.97}$ & $\mathbf{-1.07}$ \\
\quad Knowledge alignment & $\mathbf{+0.10}$ & $+0.16$ & $0.00$ & $+0.10$ \\
\quad Verification & $\mathbf{+0.24}$ & $\mathbf{+0.58}$ & $\mathbf{+0.73}$ & $\mathbf{+2.05}$ \\
\quad Selective attention & $\mathbf{+0.51}$ & $\mathbf{+0.59}$ & $\mathbf{+1.63}$ & $\mathbf{+1.31}$ \\
\quad Adaptive detail management & $\mathbf{+0.09}$ & $+0.07$ & $-0.11$ & $\mathbf{-0.52}$ \\
\quad Decomposition and integration & $\mathbf{+0.45}$ & $\mathbf{+0.49}$ & $\mathbf{+1.40}$ & $\mathbf{+0.60}$ \\
\quad Representational restructuring & $+0.08$ & $+0.17$ & $-0.37$ & $-0.08$ \\
\quad Pattern recognition & $\mathbf{+0.10}$ & $+0.08$ & $-0.19$ & $-0.21$ \\
\quad Abstraction & $+0.03$ & $-0.05$ & $\mathbf{-0.60}$ & $\mathbf{-1.10}$ \\
\quad Forward chaining & $\mathbf{+0.86}$ & $\mathbf{+0.54}$ & $\mathbf{+4.06}$ & $\mathbf{+0.97}$ \\
\quad Backward chaining & $\mathbf{+0.09}$ & $\mathbf{+0.26}$ & $+0.03$ & $\mathbf{+0.66}$ \\
\quad Backtracking & $-0.03$ & $\mathbf{+0.22}$ & $\mathbf{-0.49}$ & $\mathbf{+0.38}$ \\
\bottomrule
\end{tabular}}
\end{table}

\subsection{Change in Each Element from Well- to Ill-Structured Problems}\label{app:structure_by_element}
Table~\ref{tab:structure_by_element} reports, for each of the 28 elements, how the length-normalized presence score $s'_{t,e}$ changes from story problems (well-structured) to case analysis (ill-structured) on the matched corpus. For each element, we fit a linear mixed model of $s'_{t,e}$ with group (human or model), problem type, and their interaction as fixed effects, and random intercepts for problem and reasoner (participant or model). The human and model changes are the estimated effects of problem type within each group, and $q$ is the $p$-value of the group $\times$ problem type interaction, Benjamini--Hochberg corrected across the 28 elements.
The interaction is significant for \newthree{17} elements.
For eight of them, model traces contain less of the element on case analysis, while human traces contain a similar amount or more: the meta-cognitive controls \Metacognitive{self-awareness}, \Metacognitive{goal management}, and \Metacognitive{evaluation}, along with \Operation{verification}, \Operation{selective attention}, \Operation{decomposition and integration}, \Representation{temporal organization}, and \Operation{backtracking}.
For eight others, both groups use the element more on case analysis and models increase it more: \Invariant{productivity}, \Invariant{conceptual processing}, \Metacognitive{context awareness}, \Metacognitive{strategy selection}, \Representation{network organization}, \Operation{context alignment}, \Operation{adaptive detail management}, and \Operation{pattern recognition}.
\Invariant{Logical coherence} decreases in both groups, more in humans.
The remaining elements change by similar amounts in both groups, including the drop in step-by-step reasoning (\Operation{forward chaining}, \Representation{sequential organization}) and the rise in \Operation{abstraction} and \Operation{representational restructuring}.

\begin{table}[h]
\centering
\caption{Change from story problems to case analysis in length-normalized presence score of each element on the matched corpus, with the BH-corrected $q$ of the group $\times$ problem type interaction (bold: $q<0.05$); models drop self-monitoring elements on case analysis far more than humans do.}
\label{tab:structure_by_element}
\small
\resizebox{0.7\textwidth}{!}{%
\begin{tabular}{lrrr}
\toprule
Element & Human change & Model change & Interaction $q$ \\
\midrule
\multicolumn{4}{l}{\textit{Reasoning invariants}} \\
\quad \textbf{Logical coherence} & $\mathbf{-0.37}$ & $\mathbf{-0.23}$ & $\mathbf{0.034}$ \\
\quad Compositionality & $-0.26$ & $-0.32$ & $0.391$ \\
\quad \textbf{Productivity} & $\mathbf{+0.63}$ & $\mathbf{+0.84}$ & $\mathbf{0.001}$ \\
\quad \textbf{Conceptual processing} & $\mathbf{+0.50}$ & $\mathbf{+0.64}$ & $\mathbf{0.021}$ \\
\midrule
\multicolumn{4}{l}{\textit{Meta-cognitive controls}} \\
\quad \textbf{Self-awareness} & $\mathbf{+0.12}$ & $\mathbf{-0.16}$ & $\mathbf{6.7\times10^{-5}}$ \\
\quad \textbf{Context awareness} & $\mathbf{+0.66}$ & $\mathbf{+0.79}$ & $\mathbf{0.027}$ \\
\quad \textbf{Strategy selection} & $\mathbf{+0.52}$ & $\mathbf{+0.67}$ & $\mathbf{0.027}$ \\
\quad \textbf{Goal management} & $\mathbf{+0.08}$ & $\mathbf{-0.38}$ & $\mathbf{1.3\times10^{-14}}$ \\
\quad \textbf{Evaluation} & $\mathbf{-0.08}$ & $\mathbf{-0.72}$ & $\mathbf{1.1\times10^{-18}}$ \\
\midrule
\multicolumn{4}{l}{\textit{Reasoning representations}} \\
\quad Sequential organization & $-0.49$ & $-0.41$ & $0.229$ \\
\quad Hierarchical organization & $+0.23$ & $+0.24$ & $0.856$ \\
\quad \textbf{Network organization} & $\mathbf{+0.66}$ & $\mathbf{+0.78}$ & $\mathbf{0.017}$ \\
\quad Ordinal organization & $+0.17$ & $+0.22$ & $0.528$ \\
\quad Causal organization & $+0.30$ & $+0.32$ & $0.832$ \\
\quad \textbf{Temporal organization} & $\mathbf{+0.01}$ & $\mathbf{-0.22}$ & $\mathbf{6.2\times10^{-4}}$ \\
\quad Spatial organization & $+0.05$ & $+0.03$ & $0.536$ \\
\midrule
\multicolumn{4}{l}{\textit{Reasoning operations}} \\
\quad \textbf{Context alignment} & $\mathbf{+0.60}$ & $\mathbf{+0.80}$ & $\mathbf{2.5\times10^{-4}}$ \\
\quad Knowledge alignment & $+0.38$ & $+0.36$ & $0.733$ \\
\quad \textbf{Verification} & $\mathbf{-0.16}$ & $\mathbf{-0.76}$ & $\mathbf{2.9\times10^{-21}}$ \\
\quad \textbf{Selective attention} & $\mathbf{+0.05}$ & $\mathbf{-0.20}$ & $\mathbf{1.9\times10^{-4}}$ \\
\quad \textbf{Adaptive detail management} & $\mathbf{+0.40}$ & $\mathbf{+0.53}$ & $\mathbf{0.024}$ \\
\quad \textbf{Decomposition and integration} & $\mathbf{-0.08}$ & $\mathbf{-0.34}$ & $\mathbf{1.3\times10^{-4}}$ \\
\quad Representational restructuring & $+0.46$ & $+0.48$ & $0.771$ \\
\quad \textbf{Pattern recognition} & $\mathbf{+0.24}$ & $\mathbf{+0.42}$ & $\mathbf{0.002}$ \\
\quad Abstraction & $+0.48$ & $+0.50$ & $0.823$ \\
\quad Forward chaining & $-0.72$ & $-0.64$ & $0.293$ \\
\quad Backward chaining & $+0.24$ & $+0.14$ & $0.105$ \\
\quad \textbf{Backtracking} & $\mathbf{-0.01}$ & $\mathbf{-0.17}$ & $\mathbf{0.005}$ \\
\bottomrule
\end{tabular}}
\end{table}

\subsection{Accuracy Analysis}\label{app:accuracy_analysis}
\new{Tables~\ref{tab:accuracy_by_type} and~\ref{tab:accuracy_by_type_cont} report the accuracy of each of the 16 text reasoning models on each problem type, with the number of problems evaluated in parentheses. Problem types are ordered from well- to ill-structured as in Table~\ref{tab:model_performance}, followed by the types outside the structuredness continuum. The tables report both the mean over problem types and the overall accuracy pooled over problems. Strategic performance does not appear because no problem in the corpus is of this type.}

\begin{table}[h]
\centering
\scriptsize
\caption{\new{Accuracy by problem type and model for text reasoning problems (Part 1). Numbers in parentheses give the number of problems; \textbf{Mean over types} weights types equally and \textbf{Overall} pools all problems.}}
\label{tab:accuracy_by_type}
\resizebox{\textwidth}{!}{%
\begin{tabular}{lrrrrrrrr}
\toprule
\textbf{Problem Type} & \textbf{Qwen3-32B} & \textbf{Qwen3-14B} & \textbf{Qwen3-8B} & \textbf{R1-Qwen-32B} & \textbf{R1-Qwen-14B} & \textbf{R1-Qwen-7B} & \textbf{R1-Qwen-1.5B} & \textbf{R1-Llama-70B} \\
\midrule
Algorithmic & 78.2\% (6059) & 77.2\% (6060) & 74.4\% (6059) & 69.9\% (6061) & 65.6\% (6061) & 61.7\% (6061) & 46.2\% (6061) & 67.9\% (6061) \\
Story Problem & 84.5\% (304) & 87.2\% (304) & 85.2\% (304) & 77.6\% (304) & 78.0\% (304) & 67.4\% (304) & 57.9\% (304) & 77.6\% (304) \\
Rule-Using & 80.4\% (491) & 62.1\% (491) & 61.5\% (491) & 59.7\% (491) & 54.4\% (491) & 49.9\% (491) & 28.5\% (491) & 58.5\% (491) \\
Decision-Making & 71.9\% (89) & 78.7\% (89) & 76.4\% (89) & 57.3\% (89) & 52.8\% (89) & 37.1\% (89) & 23.6\% (89) & 65.2\% (89) \\
Troubleshooting & 81.7\% (115) & 65.2\% (115) & 65.2\% (115) & 65.2\% (115) & 54.8\% (115) & 19.1\% (115) & 10.4\% (115) & 66.1\% (115) \\
Diagnosis-Solution & 75.0\% (84) & 66.7\% (84) & 56.0\% (84) & 46.4\% (84) & 51.2\% (84) & 7.1\% (84) & 4.8\% (84) & 60.7\% (84) \\
Case Analysis & 85.2\% (149) & 80.5\% (149) & 73.2\% (149) & 49.7\% (149) & 59.1\% (149) & 20.8\% (149) & 4.7\% (149) & 74.5\% (149) \\
Design & 74.5\% (153) & 65.4\% (153) & 64.1\% (153) & 49.0\% (153) & 39.2\% (153) & 26.8\% (153) & 8.5\% (153) & 51.0\% (153) \\
Dilemma & 94.5\% (235) & 87.8\% (237) & 94.1\% (237) & 71.3\% (237) & 79.7\% (237) & 48.9\% (237) & 16.5\% (237) & 77.6\% (237) \\
Logical & 71.6\% (67) & 76.1\% (67) & 67.2\% (67) & 70.1\% (67) & 67.2\% (67) & 46.3\% (67) & 14.9\% (67) & 65.7\% (67) \\
Factual Recall & 82.4\% (2832) & 80.0\% (2832) & 78.3\% (2832) & 68.3\% (2832) & 67.7\% (2832) & 42.4\% (2832) & 22.2\% (2832) & 73.6\% (2832) \\
Creative/Expressive & 100.0\% (6) & 83.3\% (6) & 83.3\% (6) & 83.3\% (6) & 83.3\% (6) & 50.0\% (6) & 0.0\% (6) & 83.3\% (6) \\
\midrule
\textbf{Mean over types} & \textbf{81.7\%} & \textbf{75.8\%} & \textbf{73.2\%} & \textbf{64.0\%} & \textbf{62.7\%} & \textbf{39.8\%} & \textbf{19.8\%} & \textbf{68.5\%} \\
\textbf{Overall} & \textbf{80.0\%} & \textbf{77.4\%} & \textbf{75.2\%} & \textbf{68.3\%} & \textbf{65.5\%} & \textbf{53.6\%} & \textbf{36.4\%} & \textbf{69.2\%} \\
\bottomrule
\end{tabular}
}
\end{table}

\begin{table}[h]
\centering
\scriptsize
\caption{\new{Accuracy by problem type and model (Part 2), and average across all 16 models. DeepSeek-R1's ClaimSpect problems are judged by comparison with a reference answer.}}
\label{tab:accuracy_by_type_cont}
\resizebox{\textwidth}{!}{%
\begin{tabular}{lrrrrrrrr|r}
\toprule
\textbf{Problem Type} & \textbf{R1-Llama-8B} & \textbf{Olmo-3-32B} & \textbf{Olmo-3-7B} & \textbf{OpenThinker} & \textbf{s1.1-32B} & \textbf{Hermes-8B} & \textbf{DeepScaleR} & \textbf{R1-671B} & \textbf{Avg.} \\
\midrule
Algorithmic & 50.5\% (6061) & 57.8\% (6061) & 42.6\% (6061) & 72.2\% (6060) & 61.6\% (6061) & 30.5\% (6061) & 49.6\% (6061) & 81.3\% (6061) & 61.7\% \\
Story Problem & 62.8\% (304) & 68.8\% (304) & 64.1\% (304) & 79.9\% (304) & 68.4\% (304) & 48.4\% (304) & 65.1\% (304) & 85.5\% (304) & 72.4\% \\
Rule-Using & 36.7\% (491) & 59.3\% (491) & 25.7\% (491) & 75.8\% (491) & 49.5\% (491) & 33.0\% (491) & 30.8\% (491) & 85.9\% (491) & 53.2\% \\
Decision-Making & 47.2\% (89) & 66.3\% (89) & 58.4\% (89) & 74.2\% (89) & 55.1\% (89) & 37.1\% (89) & 14.6\% (89) & 80.9\% (89) & 56.0\% \\
Troubleshooting & 36.5\% (115) & 67.8\% (115) & 38.3\% (115) & 73.9\% (115) & 66.1\% (115) & 44.3\% (115) & 10.4\% (115) & 84.3\% (115) & 53.1\% \\
Diagnosis-Solution & 27.4\% (84) & 65.5\% (84) & 29.8\% (84) & 69.0\% (84) & 56.0\% (84) & 25.0\% (84) & 6.0\% (84) & 86.9\% (84) & 45.8\% \\
Case Analysis & 43.0\% (149) & 76.5\% (149) & 55.0\% (149) & 81.9\% (149) & 60.4\% (149) & 28.2\% (149) & 5.4\% (149) & 91.9\% (149) & 55.6\% \\
Design & 28.8\% (153) & 66.7\% (153) & 46.4\% (153) & 67.3\% (153) & 49.7\% (153) & 21.6\% (153) & 7.8\% (153) & 88.9\% (153) & 47.2\% \\
Dilemma & 68.8\% (237) & 89.9\% (237) & 74.3\% (237) & 92.0\% (237) & 68.4\% (237) & 47.7\% (237) & 17.3\% (237) & 87.8\% (237) & 69.8\% \\
Logical & 29.9\% (67) & 44.8\% (67) & 35.8\% (67) & 67.2\% (67) & 58.2\% (67) & 26.9\% (67) & 22.4\% (67) & 85.1\% (67) & 53.1\% \\
Factual Recall & 47.3\% (2832) & 74.0\% (2832) & 54.7\% (2832) & 80.1\% (2832) & 67.8\% (2832) & 46.5\% (2832) & 22.3\% (2832) & 88.0\% (2832) & 62.2\% \\
Creative/Expressive & 83.3\% (6) & 83.3\% (6) & 100.0\% (6) & 66.7\% (6) & 66.7\% (6) & 66.7\% (6) & 16.7\% (6) & 83.3\% (6) & 70.8\% \\
\midrule
\textbf{Mean over types} & \textbf{46.8\%} & \textbf{68.4\%} & \textbf{52.1\%} & \textbf{75.0\%} & \textbf{60.6\%} & \textbf{38.0\%} & \textbf{22.4\%} & \textbf{85.8\%} & \textbf{58.4\%} \\
\textbf{Overall} & \textbf{48.9\%} & \textbf{63.8\%} & \textbf{46.6\%} & \textbf{75.2\%} & \textbf{62.8\%} & \textbf{35.8\%} & \textbf{38.6\%} & \textbf{84.0\%} & \textbf{61.3\%} \\
\bottomrule
\end{tabular}
}
\end{table}

\subsection{\newtwo{Reasoning Topologies on the Matched Corpus}}\label{app:structures_matched}
\newtwo{Table~\ref{tab:structures_matched} lists the successful ($G^*$) and \newthree{typical} ($\hat{G}$) topologies of human and model traces on case analysis problems in the matched corpus, extracted with the procedure of Section~\ref{sec:pattern_extraction}. Each element is shown with its \newthree{presence rate} and its NPMI with correctness, computed over the traces of the same problem type and group. As in the main analysis, a trace enters the graphs through the elements present in it (presence score 2), and each topology has seven elements.}

\begin{table}[h]
\centering
\caption{\newtwo{Successful ($G^*$) and \newthree{typical} ($\hat{G}$) seven-element topologies of human and model traces on case analysis problems in the matched corpus, with elements in order of extraction, each followed by its \newthree{presence rate} and NPMI ($\rightarrow$ \texttt{NEXT}, $\parallel$ \texttt{PARALLEL}, $\supset$ \texttt{CONTAINS}; \textbf{bold}: selective attention and logical coherence). Successful topologies of both groups begin with selective attention within their first two elements and do not include forward chaining.}}
\label{tab:structures_matched}
\scriptsize
\begin{tabular}{@{}l>{\raggedright\arraybackslash}p{0.87\textwidth}@{}}
\toprule
Topology & Elements \\
\midrule
\multicolumn{2}{l}{\textit{Case analysis problems \newtwo{(ill-structured)}}} \\
\quad Human $G^*$ & \textbf{Selective attention}\,{\tiny(0.39, 0.63)} $\rightarrow$ \textbf{Logical coherence}\,{\tiny(0.54, 0.48)} $\parallel$ Compositionality\,{\tiny(0.29, 0.49)} $\parallel$ Sequential organization\,{\tiny(0.25, 0.50)} $\parallel$ Context awareness\,{\tiny(0.13, 0.40)} $\rightarrow$ Strategy selection\,{\tiny(0.20, 0.40)} $\rightarrow$ Conceptual processing\,{\tiny(0.17, 0.40)} \\
\quad Model $G^*$ & Compositionality\,{\tiny(0.74, 0.42)} $\parallel$ \textbf{Selective attention}\,{\tiny(0.65, 0.36)} $\rightarrow$ \textbf{Logical coherence}\,{\tiny(0.99, 0.04)} $\parallel$ Conceptual processing\,{\tiny(0.37, 0.29)} $\parallel$ Sequential organization\,{\tiny(0.81, 0.20)} $\supset$ Productivity\,{\tiny(0.34, 0.24)} $\rightarrow$ Goal management\,{\tiny(0.19, 0.18)} \\
\quad Human $\hat{G}$ & \textbf{Logical coherence}\,{\tiny(0.54, 0.48)} $\parallel$ \textbf{Selective attention}\,{\tiny(0.39, 0.63)} $\parallel$ Context awareness\,{\tiny(0.13, 0.40)} $\rightarrow$ Strategy selection\,{\tiny(0.20, 0.40)} $\parallel$ Compositionality\,{\tiny(0.29, 0.49)} $\parallel$ Sequential organization\,{\tiny(0.25, 0.50)} $\rightarrow$ Goal management\,{\tiny(0.11, 0.32)} \\
\quad Model $\hat{G}$ & \textbf{Logical coherence}\,{\tiny(0.99, 0.04)} $\parallel$ Sequential organization\,{\tiny(0.81, 0.20)} $\parallel$ \textbf{Selective attention}\,{\tiny(0.65, 0.36)} $\rightarrow$ Compositionality\,{\tiny(0.74, 0.42)} $\parallel$ Conceptual processing\,{\tiny(0.37, 0.29)} $\rightarrow$ Strategy selection\,{\tiny(0.29, 0.20)} $\parallel$ Productivity\,{\tiny(0.34, 0.24)} \\
\bottomrule
\end{tabular}
\end{table}

\subsection{\newthree{Test-Time Reasoning Guidance}}\label{app:guidance}
For each problem type, we extract the successful topology $G^*$ from the large corpus with the procedure of Section~\ref{sec:pattern_extraction}. \newthree{The topologies of all problem types have seven elements.}

\newthree{We then write each topology as plain text: a numbered list of its elements, followed by its edges, one per line, written as ``$A \rightarrow B$'' for \texttt{NEXT}, ``$A$ (contains) $B$'' for \texttt{CONTAINS}, and ``$A$ (parallel with) $B$'' for \texttt{PARALLEL}. This text is inserted into a fixed instruction (below), which describes the problem types, explains the three edge types, and asks for a prompt template that makes each element intuitive for the problem type, makes the relationships between the elements clear, and has a slot for the problem. We give the instruction to \texttt{Claude Sonnet 4.5}, which writes one prompt per problem type, and each problem is inserted into the prompt of its type.}

\newthree{The resulting prompts organize the elements in phases. Elements connected by \texttt{PARALLEL} edges are carried out together within a phase, a \texttt{NEXT} edge leads to the next phase, and an element under a \texttt{CONTAINS} edge is carried out within the element that contains it. Each element is described by a goal and five concrete actions for the problem type, and each prompt asks for the final answer in a fixed JSON format. The prompts range from 344 words for story problems to 1,215 words for dilemmas. Below are the instruction and the resulting prompt for dilemma problems, with the actions of all but the first element omitted.}

\begin{tcolorbox}[
    colback=gray!5!white,
    colframe=gray!75!black,
    title={\textbf{\newthree{Instruction for Writing Test-Time Reasoning Guidance}}},
    fonttitle=\bfseries,
    left=2mm,
    right=2mm,
    top=2mm,
    bottom=2mm
]
\small
We would like to prompt a model once to answer a question by utilizing this optimal reasoning structure to ``structure''/``scaffold'' its reasoning.

\vspace{0.5em}
The different problem types that we are using vary from well-defined on the top to ill-defined on the bottom (Jonassen, 2000, p. 74), as follows:
\begin{itemize}[itemsep=0pt,topsep=2pt]
    \item Logical Problems: abstract tests of reasoning that puzzle the learner
    \item Algorithmic Problems: repeating a series of steps through a procedure or formula
    \item Story Problems: story with formula or procedure embedded
    \item Rule-Using Problems: clear purpose or goal that is constrained but not restricted to a specific procedure or method
    \item Decision-Making Problems: selecting a single option from a set of alternatives based on a set of criteria
    \item Troubleshooting Problems: fault state diagnosis
    \item Case-Analysis Problems: complex, leisure-time system with multiple ill-defined goals
    \item Design Problems: vague goal statement with few constraints; requires structuring
    \item Dilemmas: situation with contradictory positions
    \item New: Factual Recall: This type is not included in the original Jonassen paper but is used to describe problems which test your ability to remember specific information, theories, definitions, or procedural knowledge directly stated in a source.
\end{itemize}

\vspace{0.5em}
Our behavior structure is a DAG, where each node is a reasoning behavior and the edges between nodes can either be: `contains', `parallel', or `next'. This means that a reasoning behavior (e.g., hierarchical organization for structuring the reasoning process and associated knowledge) may itself have a different reasoning behavior contained within it (e.g., decomposition and integration). Parallel indicates the two behaviors occur moreorless simultaneously, and `next' indicates that the behaviors occur one and another.

\vspace{0.5em}
I want you to construct a single prompt template for a given problem type and its reasoning structure. The prompt template should:
\begin{itemize}[itemsep=0pt,topsep=2pt]
    \item Briefly describe the reasoning structure easily to the model (including making each of the behaviors intuitive such that the model will understand how to execute the behavior within its reasoning for the specific problem type).
    \item Make the relationships between the behaviors clear to the model
    \item Contextualize the overall reasoning behavior structure to the given problem type
    \item It should be able to take in as input a provided question
\end{itemize}

\vspace{0.5em}
Here is my problem type: \{p\_type\}\\
Here is the optimal \{node\_num\}-node behavior structure:

\{structure\_info\}

\vspace{0.5em}
Please output your prompt template as a Python string.
\end{tcolorbox}
\begin{tcolorbox}[
    colback=gray!5!white,
    colframe=gray!75!black,
    title={\textbf{\newthree{Test-Time Reasoning Guidance: Dilemma Problems (abbreviated)}}},
    fonttitle=\bfseries,
    left=2mm,
    right=2mm,
    top=2mm,
    bottom=2mm
]
\small
You are solving a Dilemma problem. Dilemma problems require reasoning among competing choices, principles, obligations, interests, or consequences where no option may satisfy every desirable criterion. [\ldots]

\vspace{0.5em}
\textbf{Problem to Solve}\\
\{question\}

\vspace{0.5em}
\textbf{Reasoning Structure for Dilemma Problems}\\
Follow the optimized reasoning structure below, preserving genuine sequential progression while coordinating behaviors connected in parallel rather than treating them as strictly sequential steps.

\vspace{0.5em}
\textbf{Phase 1: Selective Attention}

\textit{Selective Attention.} \textbf{Goal:} Isolate the competing values, constraints, stakeholders, consequences, and uncertainties that actually create or determine the dilemma.
\begin{itemize}[itemsep=0pt,topsep=2pt]
    \item Identify the central conflict: what important goals, principles, obligations, or outcomes cannot all be fully satisfied at once.
    \item Extract the relevant options, affected parties, hard constraints, uncertainties, and potentially irreversible consequences.
    \item Separate considerations that could materially change the choice from emotionally salient or contextual details that do not affect the underlying tradeoff.
    \item Distinguish genuine conflicts from apparent conflicts that disappear once assumptions, constraints, or alternatives are clarified.
    \item Reallocate attention if later reasoning reveals that a previously secondary consideration is decisive to the dilemma.
\end{itemize}

\vspace{0.5em}
\textbf{Phase 2: Compositionality}\\
Compositionality, Productivity, Sequential Organization, and Strategy Selection operate in coordination: construct the dilemma from its constituent considerations while generating alternatives, organizing dependencies, and selecting an appropriate way to resolve the tradeoff.

\textit{Compositionality.} \textbf{Goal:} Build a structured account of the dilemma by combining its competing considerations without losing the distinct role of each one. [\ldots]

\textit{Coordinated Behavior: Productivity.} \textbf{Goal:} Generate a sufficiently broad set of defensible ways to understand or respond to the dilemma rather than remaining trapped in an unnecessarily narrow choice set. [\ldots]

\textit{Coordinated Behavior: Sequential Organization.} \textbf{Goal:} Arrange the reasoning so that prerequisite judgments and dependent decisions are handled in the correct order. [\ldots]

\textit{Coordinated Behavior: Strategy Selection.} \textbf{Goal:} Choose a decision approach suited to the type of conflict, available evidence, constraints, and stakes, and revise that approach when it fails to resolve the dilemma adequately. [\ldots]

\vspace{0.5em}
\textbf{Phase 3: Adaptive Detail Management}\\
Adaptive Detail Management and Logical Coherence operate in parallel: examine decisive details closely while repeatedly checking the global balance and consistency of the emerging resolution.

\textit{Adaptive Detail Management.} \textbf{Goal:} Adjust the level of detail so that consequential distinctions receive close analysis without allowing local details to obscure the overall dilemma. [\ldots]

\textit{Coordinated Behavior: Logical Coherence.} \textbf{Goal:} Ensure that the final treatment of the dilemma is internally consistent across assumptions, principles, evidence, comparisons, and conclusions. [\ldots]

\vspace{0.5em}
\textbf{Your Task}\\
Apply the complete reasoning structure above to the supplied dilemma. Preserve the genuine progression between sequential phases while coordinating behaviors connected in parallel.
\begin{enumerate}[itemsep=0pt,topsep=2pt]
    \item Phase 1 --- Selective Attention [\ldots]
    \item Phase 2 --- Compositionality with Productivity, Sequential Organization, and Strategy Selection [\ldots]
    \item Phase 3 --- Adaptive Detail Management with Logical Coherence [\ldots]
\end{enumerate}

\textbf{Output Format}\\
The final response must be valid JSON containing exactly one key, \texttt{final\_answer}; its string value should contain the answer to the supplied question. [\ldots]
\end{tcolorbox}

\end{document}